\documentclass[10pt,twocolumn,letterpaper]{article}

\usepackage{cvpr}           % To produce the CAMERA-READY version
\definecolor{cvprblue}{rgb}{0.21,0.49,0.74}
\usepackage[pagebackref,breaklinks,colorlinks,allcolors=cvprblue]{hyperref}
\usepackage{multirow}
\usepackage{colortbl}
\usepackage[dvipsnames]{xcolor}
\usepackage{makecell}
\usepackage{soul}
\usepackage{listings}
\usepackage[accsupp]{axessibility}  % Improves PDF readability for those with disabilities.
\def\paperID{11467} % *** Enter the Paper ID here
\def\confName{CVPR}
\def\confYear{2025}

\title{Charm: The Missing Piece in ViT fine-tuning for Image Aesthetic Assessment}

\author{Fatemeh Behrad, 
Tinne Tuytelaars, 
Johan Wagemans\\
KU Leuven University, Belgium \\
{\tt\small \{fatemeh.behrad, tinne.tuytelaars, johan.wagemans\}@kuleuven.be}
}
\begin{document}
\maketitle
\begin{abstract}
The capacity of Vision transformers (ViTs) to handle variable-sized inputs is often constrained by computational complexity and batch processing limitations. 
Consequently, ViTs are typically trained on small, fixed-size images obtained through downscaling or cropping. While reducing computational burden, these methods result in significant information loss, negatively affecting tasks like image aesthetic assessment.
We introduce \textbf{Charm}, a novel tokenization approach that preserves \textbf{C}omposition, \textbf{H}igh-resolution, \textbf{A}spect \textbf{R}atio, and \textbf{M}ulti-scale information simultaneously.
Charm prioritizes high-resolution details in specific regions while downscaling others, enabling shorter fixed-size input sequences for ViTs while incorporating essential information. Charm is designed to be compatible with pre-trained ViTs and their learned positional embeddings.
By providing multiscale input and introducing variety to input tokens, Charm improves ViT performance and generalizability for image aesthetic assessment. We avoid cropping or changing the aspect ratio to further preserve information.
Extensive experiments demonstrate significant performance improvements on various image aesthetic and quality assessment datasets (up to 8.1 \%) using a lightweight ViT backbone. Code and pre-trained models are available at \href{https://github.com/FBehrad/Charm}{GitHub}. 
\end{abstract}

% The ABSTRACT is to be in fully justified italicized text, at the top of the left-hand column, below the author and affiliation information.
% Use the word ``Abstract'' as the title, in 12-point Times, boldface type, centered relative to the column, initially capitalized.
% The abstract is to be in 10-point, single-spaced type.
% Leave two blank lines after the Abstract, then begin the main text.
% Look at previous \confName abstracts to get a feel for style and length.

\section{Introduction}
\label{sec:intro}

\begin{figure}
    \centering
    \includegraphics[width=1\linewidth]{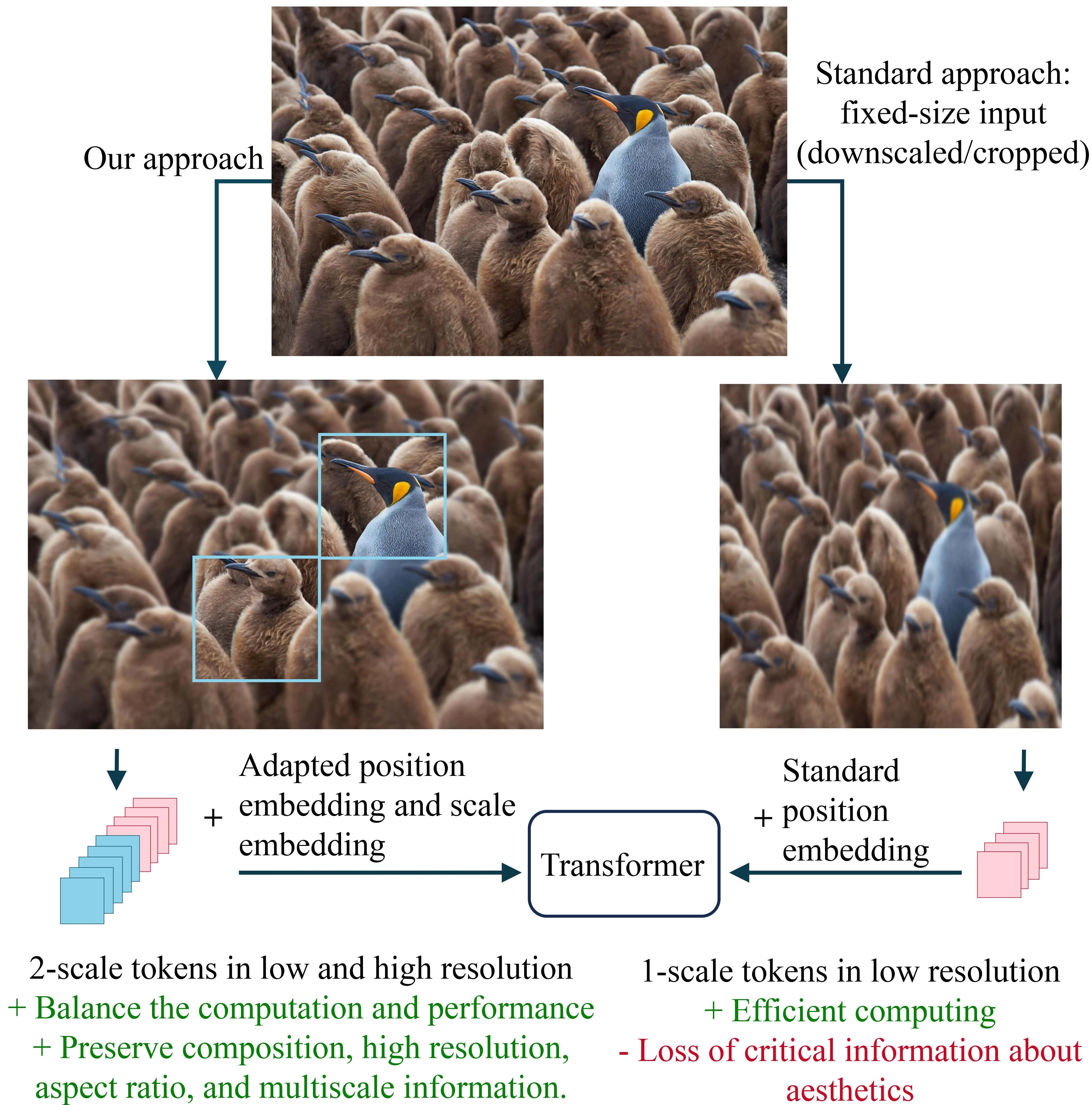}
    \caption{Traditional image preprocessing methods (right) often distort image aesthetics by rescaling and cropping images to a fixed size. To preserve critical aesthetic information, our approach (left) selectively maintains some regions in their original resolution while downscaling the others.}
    \label{fig:intro}
\end{figure}

Humans are inherently drawn to aesthetically pleasing images, making image aesthetics a vital consideration in fields like photography, advertising, e-commerce, and social media. Image aesthetic assessment (IAA) aims to develop computational models that can accurately predict human aesthetic preferences for images.
% , which are influenced by factors such as composition, image details, and aspect ratio.

Deep learning approaches applied to IAA have evolved significantly, with the Vision Transformer (ViT) architecture \cite{dosovitskiy2020image} demonstrating promising results \cite{ke2023vila, hentschel2022clip, ke2021musiq}. 
ViTs process images as sequences of non-overlapping patches, allowing them to handle images of various sizes. However, in practice, this flexibility is limited for several reasons. First, position embeddings of pre-trained ViTs are learned for a specific, fixed input size. Additionally, batch processing requires all sequences in a batch to be the same length, which is challenging with images of different dimensions. Grouping images by size or padding can help. However, grouping images may affect the learning process and convergence, as it makes batches less representative of the entire distribution, while padding can lead to inefficient resource utilization.

This brings us to another major issue when processing high-resolution images, which is the excessive computation and memory requirements. This can be particularly challenging when dealing with limited computational resources. As a result, a standard approach is to resize and downscale images to a smaller, fixed size to manage these constraints more effectively. This approach works for tasks where downscaling or cropping does not affect labels. However, in IAA, these operations can change aesthetic scores by altering the image composition and aspect ratio and losing high-resolution details (Figure \ref{fig:intro}). To unlock the full potential of ViTs in IAA, a new approach is needed to preserve critical aesthetic information while remaining computationally practical.
% ViTs can be prone to overfitting, especially when fine-tuned on small datasets, which is a common challenge in IAA. However, most data augmentation methods significantly alter images, which can be beneficial for overcoming overfitting in other tasks but can change the aesthetic score in IAA \cite{strafforello2024backflip}. 
% While ViTs theoretically handle any number of patches, considerations such as the requirement for a consistent spatial resolution in minibatch learning, memory limitations, and the computational demands associated with larger input sizes lead to the common practice of downscaling or cropping images to a smaller resolution (e.g., $224\times 224$) \cite{talebi2021learning}.

Several techniques, such as graph attention networks \cite{ghosal2022image}, adaptive fractional dilated convolution networks \cite{chen2020adaptive}, and multi-scale ViT \cite{ke2021musiq}, have been proposed to preserve image aspect ratio for IAA. Others, such as native resolution ViT \cite{dehghani2024patch}, extend beyond IAA, demonstrating the benefits of preserving aspect ratio in other tasks. However, to remain practical, existing approaches still require downscaling the entire image, which limits their effectiveness.
Fixed-size inputs can also limit the generalization of IAA models across various image sizes, which can be addressed by multi-scale inputs \cite{ke2021musiq, ke2023mret, hosu2019effective, wiedemann2023konx}. 

One possible solution to preserve high-resolution information is a learnable resizer that reduces image dimensions while minimizing information loss during downsampling \cite{talebi2021learning, cheng2020higherhrnet}.
While effective for other tasks, these methods can introduce distortions that alter the aesthetic quality of images. Another approach is to prioritize high-resolution information in some regions while downscaling less critical areas. This solution is inspired by the human brain, which processes visual information efficiently by preserving both low- and high-resolution details across the cortical hierarchy, maintaining spatial relationships, and utilizing selective attention and eye movements \cite{hochstein2002view, gilbert2013top}. Mixed-resolution \cite{ronen2023vision} and mixed-scale tokenization \cite{havtorn2023msvit} are steps in this direction. However, they still require significant downscaling or cropping for batch processing, which is not problematic for their intended task of image classification.
% In addition, mixed-resolution tokenization only supports square images and needs an extra pre-training stage to make the ViT compatible with their custom position embedding.

\textbf{Our contribution.} Existing approaches often prioritize a single factor crucial for aesthetics, neglecting others. Preserving composition, high resolution, aspect ratio, and multi-scale information simultaneously remains a challenge. We introduce \textit{Charm} to fill this gap. \textit{Charm} avoids fully downscaling images and maintains some areas in their original resolution (Figure \ref{fig:intro}). Our goal is to create a balance between computational efficiency and network performance. \textit{Charm} effectively captures aspect ratio, composition, and multiscale information.
%Additionally, our approach helps mitigate overfitting by introducing variety into the ViT's input.
Our framework-agnostic tokenization strategy avoids extensive pre-training by not modifying the pre-trained ViT architecture or its learned position embeddings.
To demonstrate the broader applicability of our approach, we evaluate it on both IAA and image quality assessment (IQA) tasks.

\section{Related Work}
\label{sec:related_work}
% MAX 2 columns
\textbf{Image aesthetic assessment}
Existing deep learning-based IAA methods often 
% suffer from limitations imposed by traditional image preprocessing techniques. These methods typically
require images to be resized or cropped to a relatively small fixed size, leading to loss of
% information loss and distortions that can degrade image aesthetics. Downscaling and cropping can alter
aspect ratio, composition, and high-resolution detail. 
Using a batch size of 1 avoids the need for image downscaling or cropping \cite{mai2016composition}. However, this can be computationally inefficient for large-scale training. Graph-based representations, such as those proposed by Ghosal et al. \cite{ghosal2022image}, can be memory-intensive for large images. Similarly, adaptive fractional dilated convolutions, introduced by Chen et al. \cite{chen2020adaptive}, can be computationally expensive and require careful batch processing.
A more recent strategy for preserving high resolution and aspect ratio is the use of learnable resizers. These methods employ auxiliary networks to reduce the dimensionality of input images while minimizing information loss during downsampling \cite{talebi2021learning, wang2022u, cheng2020higherhrnet, wang2020deep}. While effective for image classification, these approaches introduce distortions that negatively impact image aesthetics.

%, potentially hindering model performance.
Another challenge associated with fixed-size inputs is a decrease in model generalization across various image sizes. While this challenge can be mitigated by supplying multi-scale inputs to the network \cite{ke2021musiq, ke2023mret, hosu2019effective, wiedemann2023konx}, existing approaches introduce non-negligible computational costs.

\textbf{Vision transformers}
Among existing neural networks, ViTs offer the advantage of processing images in their original size. However, they are still constrained by the need for fixed-size inputs due to batch processing.
NaViT \cite{dehghani2024patch} addresses this problem by packing patches from different images into a fixed-size sequence. However, this approach can be computationally expensive as it can result in excessively long input sequences for high-resolution images.

The computational costs of ViTs are influenced by the number of input tokens they process, so downscaling images can reduce the number of tokens and improve efficiency. 
% Still, it is not suitable for IAA tasks due to its detrimental impact on image quality.
Token dropping, either randomized or structured \cite{li2024ailurus}, and token merging \cite{bolya2022token} are techniques to improve ViT efficiency. Some methods, like token fusion \cite{kim2024token}, combine these approaches to achieve even better results.
% AiluRus , for example, uses a spatial-aware density-based clustering algorithm to detect and remove less representative tokens.
% AiluRus shows that approximately 1/4 of the tokens capture most of the shape and contour information of the original images, which is adequate for dense prediction tasks. 
% However, these methods may not capture all the necessary information for IAA tasks. Therefore, for IAA, the focus must be on keeping all tokens and not discarding any information as much as possible. 

Previous studies have shown that incorporating multi-scale features can enhance ViT performance. MUSIQ \cite{ke2021musiq}, for instance, utilizes patches from different scales to capture multi-scale information. 
While effective, this approach modifies the first layers of the ViT and introduces new position embeddings, which necessitates extensive training for each ViT backbone.
FlexiViT \cite{Beyer_2023_CVPR}, a more recent approach, extracts multiscale features using random patch sizes in each training epoch.

Closer to our approach are mixed-resolution \cite{ronen2023vision} and mixed-scale tokenization \cite{havtorn2023msvit}. These methods create two-scale patches from images, preserving important regions in higher resolution and downscaling others. While effective in image classification, these approaches still require downscaling images to a relatively small fixed input size, which can negatively impact image aesthetics. Mixed-resolution ViT \cite{ronen2023vision} further limits the input format, requiring images to be square. Additionally, it replaces learned position embeddings with sinusoidal ones, necessitating additional pre-training of each ViT backbone for compatibility.

Existing methods fail to simultaneously preserve multiscale information, aspect ratio, composition, and high-resolution details. While all of these factors are crucial for accurate IAA, current methods typically prioritize computational efficiency at the expense of preserving these critical aspects. In contrast, our method successfully preserves all these factors while remaining computationally practical.
% This offers greater flexibility and applicability for processing images of any size and aspect ratio.

\textbf{Position embedding}
Transformers lack inherent mechanisms to capture the order of elements within an input sequence \cite{vaswani2017attention}. This stems from their reliance on self-attention, which is permutation-invariant – meaning the order of elements in the input does not affect the final output \cite{bello2019attention}. As a result, they require position embeddings to encode the sequence order. These position embeddings can be a deterministic function of position, typically using sinusoidal functions \cite{vaswani2017attention} or learned representations. 
Learnable position embeddings have achieved comparable performance to deterministic approaches \cite{dosovitskiy2020image}. However, both deterministic and learnable position embeddings are pre-trained for a specific sequence length. 

Relative position embeddings offer a more flexible approach by capturing the relative distance between elements \cite{shaw2018self}. This approach is well-suited for variable-length inputs. However, it necessitates significant modifications to the transformer attention mechanism.
Recent methods like hash-based spatial embedding \cite{ke2021musiq} and factorized position embedding \cite{dehghani2024patch} support variable aspect ratios and input resolutions without modifying attention modules. However, they cannot replace the position embedding of a pre-trained transformer without extensive retraining of the backbone on large-scale datasets. In contrast, our method adapts the position embedding of a pre-trained ViT model, requiring no additional pre-training and enabling efficient fine-tuning on downstream tasks with limited data.

\section{Unlock the potentials of ViTs in IAA}
\label{sec:method}
% MAX 4 columns
\begin{figure}
    \centering
    \includegraphics[width=0.8\linewidth]{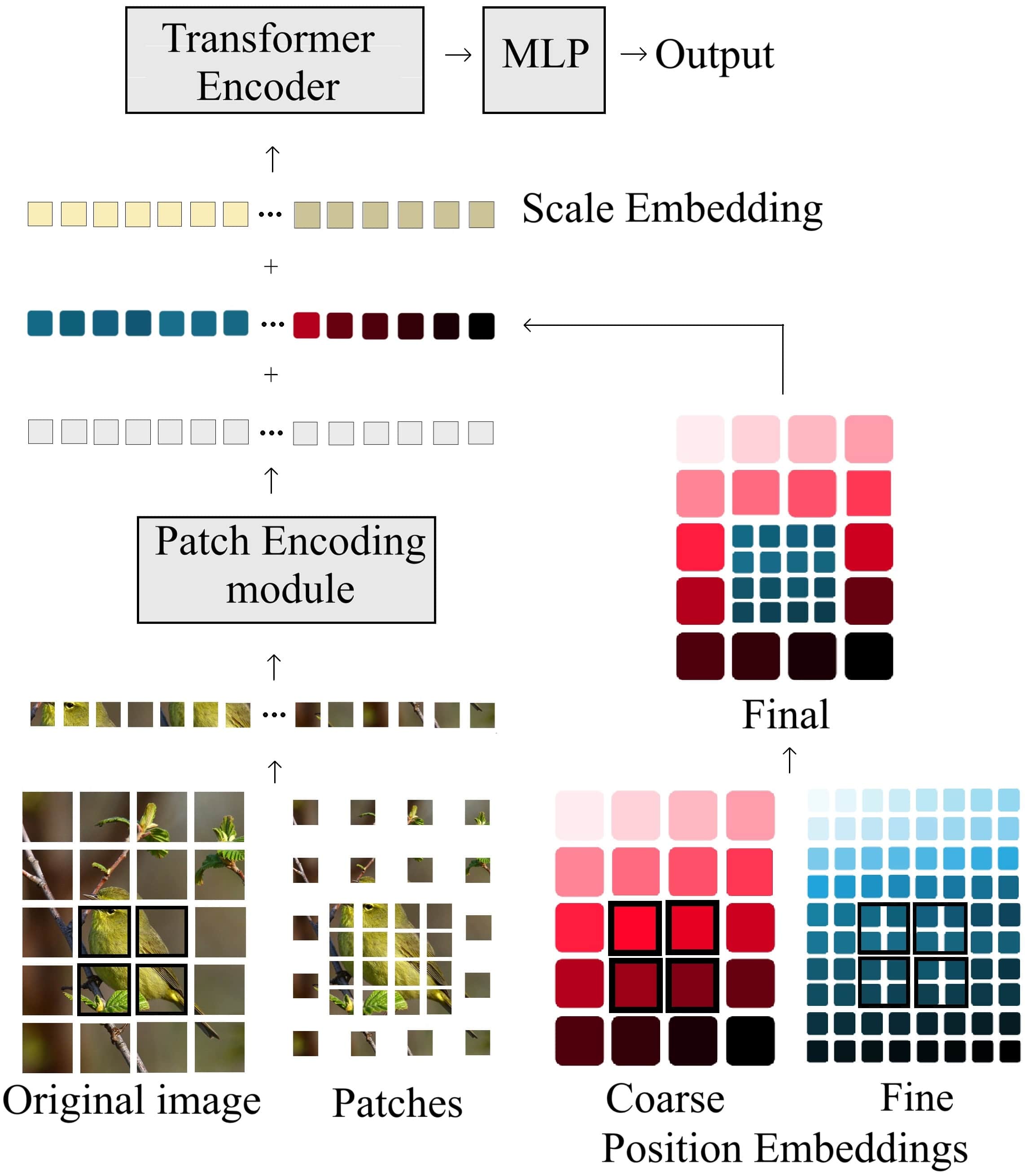}
    \caption{Our approach involves partitioning the input image into fixed-size patches and selectively preserving a subset at its original resolution while downscaling others. Position and scale embeddings are then prepared to encode spatial relationships across different scales. The remaining steps follow the standard ViT. + indicates element-wise addition.
    }
    \label{fig:architecture}
\end{figure}
In this section, we introduce \textit{Charm}, a novel tokenization approach designed to enhance the performance of standard ViTs in image aesthetic and quality assessment tasks. Figure \ref{fig:architecture} provides an overview of our proposed approach.

We initially partition the input image into fixed-size patches, preserving a selection of patches at their original resolution while downscaling others. This approach enables the model to capture both fine-grained details and broader contextual information. Subsequently, position embeddings are prepared to encode the spatial relationships between patches across different resolutions and within the same resolution. Following previous studies \cite{ke2021musiq}, we also incorporate scale embeddings to aid the model in distinguishing information from different scales. 

After encoding the input patches into a sequence of tokens using the patch encoding module, we add position and scale embeddings. A learnable "classification token" (CLS) is prepended, and all tokens are processed through the transformer encoder. The final output is obtained by feeding the CLS token state at the output of the transformer through a fully connected layer. 
Our approach is compatible with various ViT architectures, as we do not modify the core ViT network structure.

\subsection{Preliminaries}
An image $x \in \mathbb{R}^{h \times w \times c}$, where $(h, w, c)$ represents the height, width, and number of channels, respectively, undergoes tokenization in a standard ViT. This process divides the image into a sequence of $s$ patches $x_i \in \mathbb{R}^{p \times p \times c}$, where $i \in (1, \dots, s)$ and $p$ represents the patch size. The sequence length $s = \left \lfloor \frac{h}{p} \right \rfloor \cdot \left \lfloor \frac{w}{p} \right  \rfloor$ determines the number of patches after tokenization, controlling the ViT's computational complexity. Given that the patch size $p$ remains fixed for a given ViT, the number of patches depends on the input image resolution. Consequently, high-resolution images increase computational complexity.
% Preserving some patches in high resolution while downsizing the others results in a shorter sequence of tokens ($s$) while preserving some high-resolution information.

\subsection{Charm tokenization}
\label{tokenization}

\subsubsection{Tokenization}
\label{3.2.1}
Batch processing necessitates a consistent spatial resolution within each minibatch, leading to a fixed input length ($l$) for the ViT. Processing high-resolution images requires a large $l$, which increases memory consumption and computational demands during training. Our method addresses this by strategically reducing $l$ while preserving aesthetic-related information (Figure \ref{fig:architecture}).
Given a fixed patch size ($p$), we can consider two scenarios. For low-resolution images ($s \leq l$), we follow the standard ViT approach, tokenizing the entire image with the patch size of $p$. 
For high-resolution images ($s > l$), we keep some parts in high resolution while downscaling other areas. 

When dealing with 2 scales, we initially tokenize the image using a patch size of $p' = p \times n$, where $n \geq 2$. 
Subsequently, we select high-resolution patches and further tokenize them using a patch size of $p$ while the remaining patches are downscaled to $p$. The downscaling factor for both height and width is $f = 1/n$.
Finally, we arrange all high-resolution patches at the beginning, followed by low-resolution ones to form the final input for the transformer encoder.  This ordering is feasible due to the permutation invariance of ViTs \cite{bello2019attention}, allowing us to avoid unnecessary implementation loops.  
To ensure a consistent input length ($l$), padding or random token dropping is employed as needed. Additional details on the 3-scale version are available in the Appendix 7.
 
% , where patch position information is learned through position embeddings. 
% A detailed description of the algorithm for this process can be found in Appendix ?." 

\subsubsection{Patch Importance}
Several strategies can be employed to determine which patches to retain at high resolution. The main goal is to preserve visually interesting areas and sharp details in high resolution. 
To identify these regions, we use frequency, gradient, entropy, and saliency maps. 
% We randomly select patches within areas exhibiting high frequency, entropy, or gradient, as well as salient regions. 
We compare them with a random selection of patches in each training epoch. More details on patch selection strategies, including illustrative examples, can be found in Figure 8 of Appendix 8.

\subsubsection{Position Embedding}
ViTs lack inherent knowledge of spatial relationships between patches. Position embeddings address this by injecting learned position information into each patch embedding, allowing the self-attention mechanism to capture how different parts of the image relate. Replacing the standard position embeddings of ViTs with custom ones \cite{ronen2023vision} can be expensive, requiring extensive pre-training on large datasets. It also introduces compatibility issues when used with different pre-trained ViTs. To address these issues, our approach adapts the pre-trained position embeddings from the standard ViT models. 

For low-resolution images, we follow the standard ViT approach of resizing pre-trained position embeddings via bilinear interpolation. However, for high-resolution images, we create two position embeddings using bilinear interpolation based on the different patch sizes ($p$ and $p'$) employed during tokenization (Figure \ref{fig:architecture}). One position embedding retains the spatial relationships between low-resolution patches, while the other captures the finer spatial relationships within the high-resolution patches. These position embeddings, high and low resolution, are compatible because they are derived from the same pre-trained embedding.
Similar to standard ViTs, each token from the output of the patch encoding module is then summed with its corresponding position embedding. 

Unlike existing methods \cite{havtorn2023msvit, ronen2023vision}, our approach does not have a fixed image size, so position embeddings vary for each image. 
To avoid recalculating position embeddings in each epoch, we prepare them in the data loaders, keeping them frozen during fine-tuning. This aligns with the standard practice of freezing the initial layers (including position embeddings) and fine-tuning only the final layers.

\subsubsection{Scale Embedding}
Our tokenization approach results in multiscale tokens. Following previous studies \cite{ke2021musiq}, to help ViT distinguish between patches from different scales and capture their relationships, we incorporate learnable scale embeddings. 
These scale embeddings are randomly initialized and added element-wise to all patch embeddings (Figure \ref{fig:architecture}).

\subsection{Fine-tuning}
% Our method does not alter the model's architecture or learned position embeddings, eliminating the need for additional pre-training. 
Our method requires no additional pre-training.
We process images in their original resolution without resizing or random cropping during fine-tuning. We avoid normalizing the images as normalization can introduce distortions that affect the overall aesthetic/quality score (Table 9 in Appendix 9).
Our data augmentation is limited to random horizontal flipping and random rotations (90, 180, 270 degrees) with a 50\% probability, as they do not result in loss of composition or aspect ratio. While these methods might change the aesthetics for humans, they have been shown to improve the performance of ViTs (Appendix  10). During evaluation, data augmentation is completely omitted.
We employ the L1 loss function for predicting the overall aesthetic/quality score and the Earth Mover's Distance (EMD) loss \cite{rubner1998metric} for predicting the aesthetic score distribution.

\section{Experiments}
\begin{figure}
    \centering
    \includegraphics[width=1\linewidth]{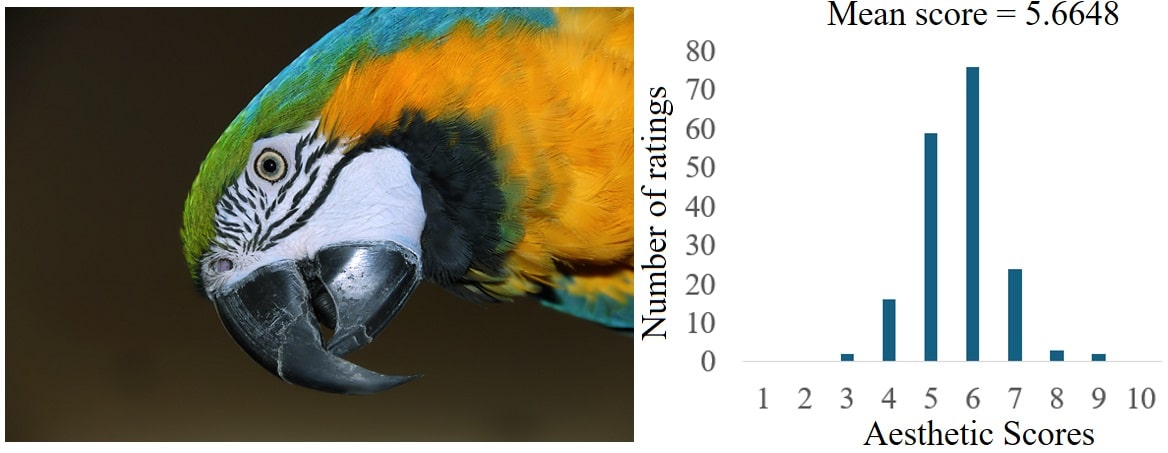}
    \caption{An image with its aesthetic score distribution and mean aesthetic score, from the AVA dataset.}
    \label{fig:example}
\end{figure}

\subsection{Datasets}
We evaluate our models on various IAA datasets, including AVA \cite{6247954}, PARA \cite{yang2022personalized}, TAD66k \cite{he2022rethinking}, AADB \cite{kong2016photo}, BAID \cite {yi2023towards}, and two IQA datasets, SPAQ \cite{fang2020perceptual} and KonIQ10k \cite{hosu2020koniq}. We only use the images and their corresponding ratings, excluding any additional information provided in the datasets.
The AVA and PARA datasets associate each image with an aesthetic score distribution, while other datasets associate each image with an overall aesthetic/quality score (Figure \ref{fig:example}). As the SPAQ and KonIQ10k datasets lack an official train-test split, we repeat the experiments for 5 different splits and report the median performance (as done in \cite{shin2024blind}).
% AVA, the largest and most popular IAA dataset, contains 235k training and 20k test images. Each image is associated with an aesthetic score distribution ranging from 1 to 10.
% PARA is primarily designed for personalized IAA and comprises 28k training images and 3k test images. Each image is annotated with an aesthetic score distribution ranging from 1 to 5 (with 0.5 intervals).
% TAD66k contains 52k training and 14k test images, categorized into 47 themes with an overall aesthetic score from 1 to 10.
% AADB provides 8.5k training, 500 validation, and 1k test images. Each image has a normalized overall aesthetic score (0-1) and annotations for 11 aesthetic attributes ranging from -1 (low) to 1 (high). 
% BAID is the first artistic image aesthetic dataset, including 54k training and 6.4k test images with overall aesthetic scores between 0 and 10.  
% The SPAQ dataset consists of 11,125 pictures taken by 66 smartphones with an overall quality score between 0 and 100. As this dataset lacks an official train-test split, we repeat training and testing for 10 different splits and report the median performance (as done in...).

The AVA dataset remains prominent due to its large number of images, allowing for effective training of large models. Research utilizing smaller datasets is relatively limited, and most existing approaches leverage the additional information provided in these datasets. However, considering various datasets with different resolution ranges allows us to validate the efficiency of our approach in various scenarios.
For more details on the resolution distribution of each dataset, see Figure 13 in Appendix 11.

\subsection{Implementation details}

Saliency maps for patch selection are generated using the SAM-HQ model \cite{ke2024segment} with the text prompt of ``salient object'' (Appendix 8). 
For gradient, frequency, and entropy calculations, we use the Sobel filter, fast Fourier transform, and Shannon entropy, respectively.

In all experiments, we fine-tune the entire ViT models,
% the entire ViT-small \cite{dosovitskiy2020image} and Dinov2-small models \cite{oquab2023dinov2},
while keeping their position embeddings frozen. 
Also, we use scale embeddings alongside position embeddings to enable \textit{Charm} capture relationships between tokens both within and across scales (see Appendix 12).
The patch size ($p$) and patch stride are chosen to match the filter size and stride of the patch encoding modules within these models (16 in ViT-small and 14 in Dinov2-small), ensuring compatibility and replicating the models' internal tokenization process.
% As we do not change the architecture of the pre-trained backbones, we preserve their original number of trainable parameters.

% For all datasets except PARA and SPAQ, fine-tuning is performed on full-resolution images. Due to computational limitations, images in PARA and SPAQ are downscaled to a maximum edge of 1024 while preserving their aspect ratio.
The maximum number of tokens ($l$) is set to 512 for AVA, 768 for TAD66k, and 1024 for other datasets. These values are chosen based on the image resolution in the datasets (Appendix 13).
For all experiments, we use a factor size of 0.5. % as a higher number of scales will not result in further improvement (Appendix ?).

The AdamW optimizer \cite{loshchilov2018fixing} is employed for fine-tuning. We use an initial learning rate of 1e-5 for all datasets except BAID (1e-6). We gradually reduce the learning rate during training using the cosine annealing learning rate scheduler.
The batch size is 64 for AVA, 32 for TAD66k, and 16 for the remaining datasets. Early stopping is also implemented to prevent overfitting. 
% For detailed hyperparameters, please refer to Appendix ?.
Fine-tuning is conducted on a single NVIDIA RTX A4500 GPU.
Model performance is evaluated using three key metrics: Spearman rank-ordered correlation coefficient (SRCC), Pearson linear correlation coefficient (PLCC), and classification accuracy (ACC).

% We use a batch size of 64 with $l$ = 512 and $f=2$ for the AVA dataset, resulting in a training time per epoch of around 50 minutes. Because of the larger average image sizes in TAD66k, patchification leads to a higher number of patches ($s$). To accommodate this, we employ a batch size of 32 with $l=768$ and $f=2$ for TAD66k, leading to a training time of approximately 20 minutes per epoch. Despite requiring a potentially higher $l$ due to its high resolution compared to other datasets, AADB achieved optimal performance with a surprisingly low sequence length of 512 and a batch size of 64. This might be attributed to AADB's smaller number of images, where longer sequences could lead to overfitting (details in Appendix \ref{appendix:optimal_sequenth_length}). Training on AADB took 5 minutes per epoch.
\begin{table}
\small
    \centering
    \begin{tabular}{m{4em}m{1cm}m{4em}m{4em}m{4em}} 
        \hline
        Dataset & Charm & PLCC & SRCC & ACC \\
        \hline
        \multicolumn{5}{c}{\cellcolor{gray!25}IAA} \\
        \hline
        \multirow {2}{*}{AVA} & - & 0.734 & 0.732 & 0.808 \\ \cline{2-5}
        & \checkmark & 0.779 \color{NavyBlue}{($\uparrow4.5\%$)}
        & 0.777 \color{NavyBlue}{($\uparrow4.5\%$)} & 0.826 \color{NavyBlue}{($\uparrow1.8\%$)} \\
        % \cline{2-5} 
        % & - * & 0.687 & 0.679 & 0.794 \\ \cline{2-5}
        % & \checkmark * & 0.762 \color{NavyBlue}{($\uparrow7.5\%$)} & 0.760 \color{NavyBlue}{($\uparrow8.1\%$)} & 0.827 \color{NavyBlue}{($\uparrow3.3\%$)}\\
        \hline
        \multirow {2}{*}{AADB}& - & 0.695 & 0.682 & 0.754\\ \cline{2-5} & \checkmark & 0.767 \color{NavyBlue}{($\uparrow7.2\%$)} & 0.754 \color{NavyBlue}{($\uparrow7.2\%$)} & 0.767 \color{NavyBlue}{($\uparrow1.3\%$)} \\ 
        \hline
        \multirow {2}{*}{TAD66k}& - & 0.429 & 0.401 & 0.646 \\ \cline{2-5} & \checkmark & 0.488 \color{NavyBlue}{($\uparrow5.9\%$)} & 0.458 \color{NavyBlue}{($\uparrow5.7\%$)} & 0.794 \color{NavyBlue}{($\uparrow14.8\%$)} \\
        \hline
        \multirow {2}{*}{PARA}& - & 0.904 & 0.855 & 0.863 \\ \cline{2-5} & \checkmark &      
        0.938 \color{NavyBlue}{($\uparrow3.4\%$)} & 0.905 \color{NavyBlue}{($\uparrow5\%$)} & 0.892 \color{NavyBlue}{($\uparrow2.9\%$)} \\
        \hline
        \multirow {2}{*}{BAID}& - & 0.428 & 0.342 & 0.750 \\ \cline{2-5} & \checkmark & 0.439 \color{NavyBlue}{($\uparrow1.1\%$)} & 0.368 \color{NavyBlue}{($\uparrow2.6\%$)} & 0.763 \color{NavyBlue}{($\uparrow1.3\%$)} \\ \hline
        \multicolumn{5}{c}{\cellcolor{gray!25}IQA} \\ \hline
        \multirow{2}{*}{SPAQ} & - & 0.911 & 0.907 & 0.907 \\ \cline{2-5} & \checkmark & 0.919 \color{NavyBlue}{($\uparrow0.8\%$)} & 0.915 \color{NavyBlue}{($\uparrow0.8\%$)} & 0.917 \color{NavyBlue}{($\uparrow1.0\%$)} \\ 
        \hline
        \multirow{2}{*}{KonIQ10k} & - & 0.896 & 0.868 & 0.938 \\ \cline{2-5}  & \checkmark & 0.944  \color{NavyBlue}{($\uparrow 4.8\%$)} & 0.93 \color{NavyBlue}{($\uparrow6.2\%$)} & 0.954  \color{NavyBlue}{($\uparrow1.6\%$)} \\ 
        \hline
    \end{tabular}
    \caption{Performance improvement across different IAA and IQA datasets by replacing the standard tokenization of Dinov2-small with \textit{Charm} (with a random patch selection strategy).}
    \label{tab:improvement}
\end{table}

\begin{table*}
\small
    \centering
    \begin{tabular}
    {lcccccc}
        \hline
        Model &AR&HR&MS&PLCC&SRCC&ACC\\
        \hline
        Ghosal et al.\cite{ghosal2022image} & \checkmark & - & -   & 0.764 & 0.762 & - \\
        Chen et al. \cite{chen2020adaptive} & \checkmark & - & - & 0.671 & 0.649 & \textbf{0.832} \\
        Dinov2-Small (+ Padding) * & \checkmark & \checkmark & - & 0.709 & 0.703 & 0.801 \\
        MUSIQ \cite{ke2021musiq} & \checkmark & - & \checkmark & 0.738 & 0.726 & 0.815 \\
        FlexiViT \cite{Beyer_2023_CVPR}* & - & - & \checkmark &  0.737 & 0.735 & 0.812 \\
        Swin \cite{liu2021swin}* & - & - & \checkmark &  0.748 & 0.751 & 0.816 \\
        % Hentschel et al.\cite{hentschel2022clip} & - & - & - & \checkmark & 0.741 & 0.731 & 0.816 \\
        % VILA \cite{ke2023vila} & - & - & - & \checkmark & \underline{0.774} & \underline{0.774} & - \\
        Dinov2-Small (+ Muller \cite{tu2023muller}) * & - & \checkmark & - & 0.682 & 0.675 & 0.794 \\
        \hline
        % ViT-Small \cite{dosovitskiy2020image} & - & - & - & - & 0.687 & 0.679 & 0.794 \\
        ViT-small (+ Charm) * & \checkmark & \checkmark & \checkmark & 0.762 %\textcolor{red}{($\uparrow7.5\%$)}
        & 0.760 
        %\textcolor{red}{($\uparrow8.1\%$)}
        & 0.827
        %\textcolor{red}{($\uparrow3.3\%$)}
        \\
        % Dinov2-Small \cite{oquab2023dinov2} & - & - & - & -  & 0.710 & 0.706 & 0.802 \\
        Dinov2-small (+ Charm) * & \checkmark & \checkmark & \checkmark & \underline{0.779}
        % \textcolor{red}{($\uparrow6.9\%$)}
        & \underline{0.777}
        % \textcolor{red}{($\uparrow7.1\%$)}
        & 0.826 \\
        % \textcolor{red}{($\uparrow2.4\%$)} 
        Dinov2-large (+ Charm) * & \checkmark & \checkmark & \checkmark & \textbf{0.783} & \textbf{0.781} & \underline{0.828}
        \\ \hline
    \end{tabular}
    \caption{Results on the AVA dataset. 
    \textbf{Bold} and \underline{underlined} numbers indicate the best and the second-best results, respectively. AR, HR, and MS represent preserving aspect ratio, high-resolution details, and multi-scale information, respectively. * indicates models that we have fine-tuned on the AVA dataset. Appendix 14 shows the performance of the standard ViT-small, Dinov2-small, and Dinov2-large.}
    \label{tab:sota}
\end{table*}

\subsection{Comparing with existing methods}
We benchmark \textit{Charm} with various ViT backbones on different IAA and IQA datasets. 
We first fine-tune models on the datasets to create a baseline. We then replace their original patch encoding modules with two-scale \textit{Charm} tokenization and fine-tune the entire models. We also considered a three-scale version of \textit{Charm} but did not observe additional performance gains (Appendix 7). Therefore, we focus on the two-scale version throughout the paper.

The results (Table \ref{tab:improvement} and Table 13 in Appendix 14) demonstrate that by simply replacing the patch encoding module of existing ViT models with \textit{Charm}, we can achieve significant performance improvements, up to 7.5\% in PLCC, 8.1\% in SRCC, and 14.8\% in classification accuracy across various datasets.
Among datasets, PARA and SPAQ datasets include extra high-resolution images (e.g., 2k by 3k pixels). Our experiments (Appendix 15) reveal a threshold for performance improvement by preserving high-resolution information. Beyond this threshold, further performance gains are limited. Therefore, to decrease computational costs, we downscale images in PARA and SPAQ to a maximum edge of 1024.

As shown in Table \ref{tab:sota}, our approach outperforms existing methods in terms of preserving resolution, aspect ratio, and multiscale information. We compare \textit{Charm} with the simple approach of adding padding to ensure consistent input shapes within each batch while preserving the aspect ratio. However, this padding proves detrimental to IAA and significantly increases computational costs (Appendix 16). Also, we observe that Muller \cite{tu2023muller}, a recent approach focusing on preserving high-resolution information, can introduce noise and distortions to the image, negatively affecting aesthetics (Appendix 16). 

Recent studies have demonstrated the effectiveness of incorporating additional modalities, such as attributes (\cite{10054147, 10433197, 10502339}) or text captions (\cite{sheng2023aesclip, 10115482, zhu2024emotion, li2024towards}), to improve IAA model performance (Table 15 in Appendix 17). Other studies have concentrated on other research directions \cite{shi2024improving, he2023eat}, like addressing the long tails of distributions in IAA datasets \cite{liuelta}.
While the main contributions of these studies are orthogonal to \textit{Charm}’s, integrating \textit{Charm} with them further enhances their performance. For example, incorporating \textit{Charm} into ELTA \cite{liuelta} with a Dinov2-large backbone improves its PLCC, SRCC, and ACC on the AVA dataset from 0.742, 0.742, and 0.811 to 0.787, 0.786, and 0.829, respectively.
Also, our approach demonstrates strong performance in IQA, with a slight performance gap of less than \textbf{1\%} in SPAQ and \textbf{0.4\%} in KonIQ10k compared to state-of-the-art methods \cite{shin2024blind, ke2021musiq, chen2024topiq} (Table 16 in Appendix 17).

\begin{figure}
    \centering
    \includegraphics[width=1\linewidth]{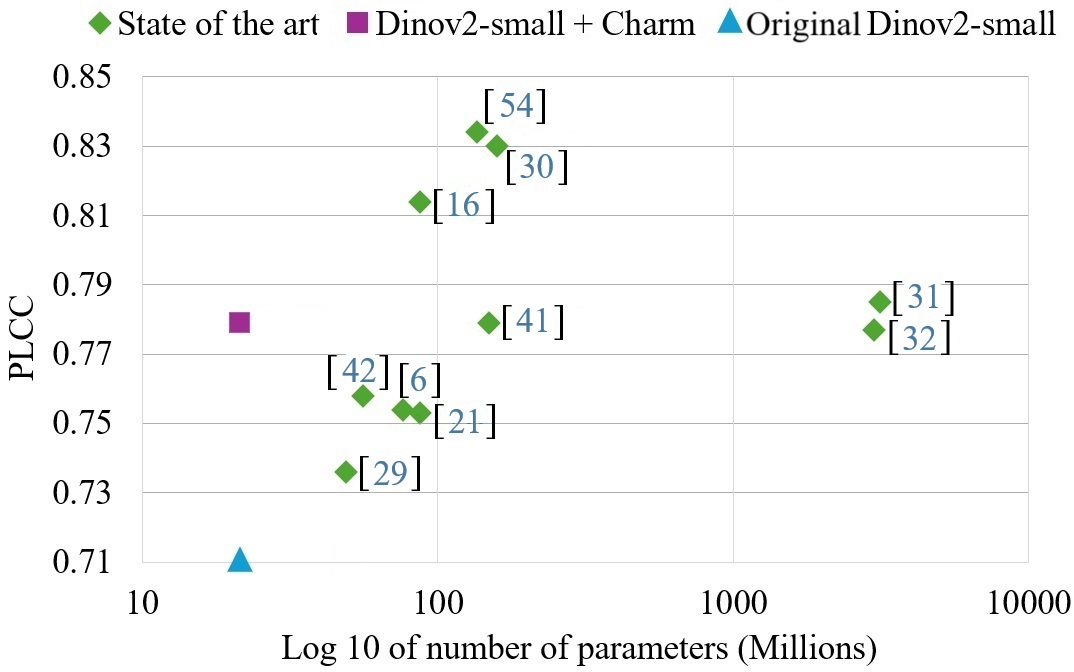}
    \caption{Comparison of state-of-the-art IAA models' performance on the AVA dataset vs. their number of parameters. Our approach achieves comparable performance without using additional modalities and with significantly fewer parameters. To account for the significant differences in the number of parameters, we use a logarithmic scale on the x-axis. For a detailed comparison of Dinov2-small + \textit{Charm} with state-of-the-art models on all datasets, see Appendix 17.}
    % Why logaritimic scale? Because: https://www.indeed.com/career-advice/career-development/logarithmic-scale
    \label{fig:params}
\end{figure}

Model size is another crucial factor, as larger models often excel in complex tasks like IAA. In this paper, we demonstrate that \textit{Charm} is highly effective even with lightweight ViTs, which can be used for applications with limited computational resources. As shown in Figure \ref{fig:params}, Dinov2-small + \textit{Charm} achieves comparable performance to state-of-the-art models while being significantly smaller.
Nevertheless, \textit{Charm} can serve as a plug-in for larger, more expensive models if higher performance is desired and computational power allows. For instance, plugging \textit{Charm} into ViT-base instead of Dinov2-small results in PLCC of \textbf{0.515}, SRCC of \textbf{0.467}, and ACC of \textbf{0.779} after only 9 epochs on the BAID dataset. 

% While Charm can be integrated with other models for potential performance gains, computational limitations and the unavailability of code for many existing studies prevented us from conducting such integrations in this research.

\subsection{Ablation study}
\subsubsection{The importance of preserving aspect ratio and high-resolution details}
\label{other_factors}
Current approaches usually distort images by resizing them to a square shape, disregarding the significance of maintaining their original size and aspect ratio. We argue that such preprocessing can be detrimental to IAA. To validate this claim, we compare the performance of Dinov2-small using different tokenization approaches. Table \ref{tab:tokenization} presents the results of these experiments.  
Except for the standard tokenization (described in Appendix 9), all other settings involve multiscale inputs and maintain composition by avoiding image cropping.

We initially downscale images to 256 x 256 and then preserve some areas at that resolution while downscaling others by a factor of 0.5 (MS in Table \ref{tab:tokenization}). While this approach has been effective in image classification (\cite{havtorn2023msvit} and \cite{ronen2023vision}), it decreases performance in our IAA task. This is because the initial downscaling to a fixed size results in significant information loss critical for IAA, and further downscaling worsens the issue. 

We then experiment with keeping the image in its original resolution while changing the aspect ratio by resizing the image based on the maximum edge to create a square image (HR + MS in Table \ref{tab:tokenization}). This results in a nearly 3\% increase in PLCC and SRCC compared to the standard approach. 
Finally, we use \textit{Charm}, which preserves both aspect ratio and high-resolution details, resulting in the best performance. This analysis highlights the importance of preserving both aspect ratio and original resolution in IAA tasks. 

We also examined the impact of different numbers of tokens across tokenization approaches in Table \ref{tab:tokenization} but found no differences in the results (see Appendix 18).
\begin{table}
\small
    \centering
    \begin{tabular}{lccc}
    \hline
        Tokenization & PLCC & SRCC & ACC \\
        \hline
        Standard & 0.734 & 0.732 & 0.808 \\
        MS & 0.710 & 0.705 & 0.801 \\
        HR + MS & 0.763 & 0.760 & 0.820 \\
        Charm (AR + HR + MS) & \textbf{0.779} & \textbf{0.777} & \textbf{0.826}  \\ \hline
    \end{tabular}
    \caption{Performance of Dinov2-small on the AVA dataset with four different tokenization strategies. MS, HR, and AR represent preserving multiscale information, high-resolution details, and aspect ratio, respectively.}
    \label{tab:tokenization}

\vspace{-1em}
\end{table} 

\subsubsection{Comparison of patch selection strategies}
To evaluate the importance of patches, we fine-tune Dinov2-small on the AADB dataset with different patch selection strategies. Table \ref{tab:patch_selection} demonstrates that all strategies outperform the standard ViT approach \cite{dosovitskiy2020image}. This suggests that factors beyond preserving high-resolution details also contribute to performance improvements (Section \ref{other_factors}).

Among all methods, the random selection of patches in each epoch stands out, likely for two reasons. First, ViTs are prone to overfitting, especially when fine-tuned on small datasets, a common challenge in IAA. Random patch selection in each epoch addresses this by introducing more variety to the ViT input and acting as a form of data augmentation. Second, random sampling can better capture the subjective and often unpredictable nature of beauty, providing a more representative view of the entire image.
Following random selection, frequency, gradient, and entropy-based methods stand out. These methods still have a certain level of randomness, as their fully deterministic version results in highly correlated selected patches (Appendix 8).

% We observed that random selection of patches in the first epoch and keeping them fixed in the training process performs better than the baseline but worse than any other patch selection strategies (Table \ref{tab:random_fixed} in Appendix \ref{appendix: patch_selection}). 

ViTs, originally trained for object classification, have an inherent bias towards salient objects. However, IAA requires attention to both foreground and background \cite{he2023eat}, which explains why the saliency-based approach performs less effectively.
For a more comprehensive analysis of patch selection strategies please refer to Appendix 8.

\begin{table}
\small
    \centering
    \begin{tabular}{lccc}
    \hline
        Patch selection & PLCC & SRCC & ACC \\
        \hline
        Random & \textbf{0.767} & \textbf{0.754} & \textbf{0.767} \\
        Entropy & 0.726 & 0.714 & 0.761\\
        Frequency & \underline{0.756} & \underline{0.747} & 0.761\\
        Gradient & 0.734 & 0.721 & \underline{0.766}\\
        Saliency & 0.751 & 0.738 & 0.756\\ 
        Standard & 0.695 & 0.682 & 0.754 \\
        \hline
    \end{tabular}
    \caption{Performance of Dinov2-small on the AADB dataset using \textit{Charm} with different patch selection strategies. \textbf{Bold} and \underline{underlined} values highlight the best and second-best results.}
    \label{tab:patch_selection}
\end{table}

\subsubsection{The effect of Charm on convergence speed}
\begin{figure}
    \centering
    \includegraphics[width=1\linewidth]{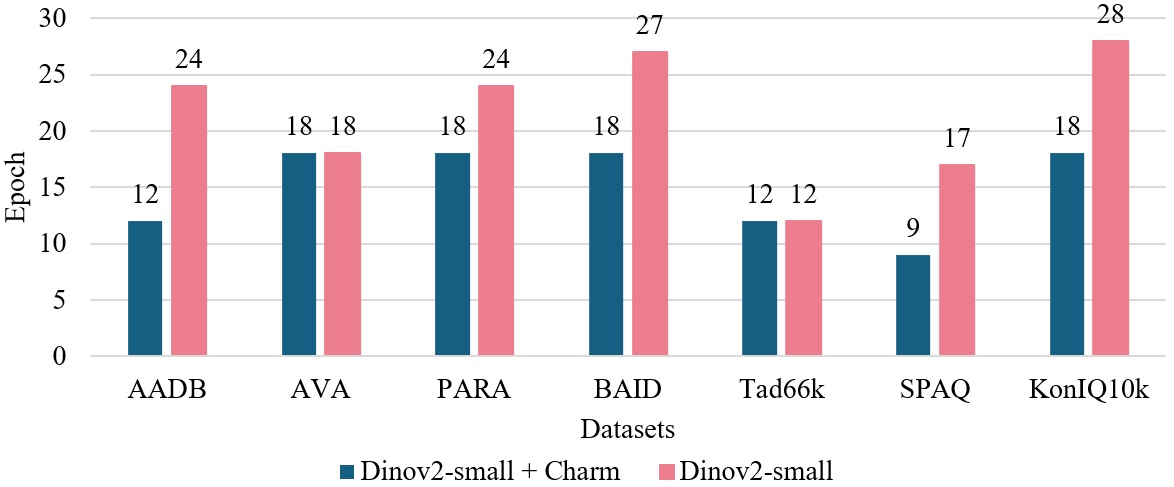}
    \caption{The epoch at which each model achieves its highest validation performance across different datasets. \textit{Charm} generally leads to faster convergence. The average number of epochs over 5 runs is reported for SPAQ and KonIQ10k. }
    \label{fig:convergence}
\end{figure}

We fine-tune all models until convergence, meaning their performance on the validation set stabilizes. Figure \ref{fig:convergence} shows the epoch at which each model achieves its best validation performance. As shown in this figure, \textit{Charm} consistently accelerates the convergence of Dinov2-small across different datasets.
Notably, the most significant reduction (50\%) in training epochs occurs on AADB, our smallest dataset, likely due to \textit{Charm}'s ability to mitigate overfitting through random patch selection (see Appendix 8). On our largest datasets, AVA and TaD66k, both standard and \textit{Charm} tokenization converge at similar epochs.

\begin{table*}
\small
\centering
    \begin{tabular}{lllllll}
        \hline
        Model & Input size& Charm & \#tokens & ms & GMACs & MB \\
        \hline
        \multirow {2}{*}{\makecell{Dinov2\\-small}} &  224 x 224 & - & 256 & \textbf{5.7} &\textbf{ 6.11} & \textbf{202.9 }\\ 
        \cline{2-7}  & 
        \multirow {3}{*}{640 x 640} & - & 2070 & 32.8 & 84.01 & 2091.8
        \\ & & \checkmark & 2-scale:512 & \underline{7.3}  \color{NavyBlue}{($\downarrow77.7\%$)}& \underline{13.46} \color{NavyBlue}{($\downarrow84\%$)} & \underline{346.0}
         \color{NavyBlue}{($\downarrow83.5\%$)}
         \\ && \checkmark & 3-scale:700 & 9.3 \color{NavyBlue}{($\downarrow71.6\%$)} & 19.60 \color{NavyBlue}{($\downarrow76.7\%$)} & 494.3 \color{NavyBlue}{($\downarrow76.4\%$)}\\ 
         \hline
    \end{tabular}
    \caption{Dinov2-small inference cost breakdown for processing one single image: number of tokens (\#tokens) based on varying input sizes, runtime in milliseconds (ms), Giga multiply accumulation (GMACs), and GPU memory in Megabytes (MB). \textbf{Bold} and \underline{underlined} values highlight the most and second-most computationally efficient configurations.
    Percentages indicate the reduction in computational cost compared to processing the image in its original size. Similar results for ViT-small are presented in Table 17 in the Appendix.}
    \label{tab:model_costs}
\end{table*}

\subsubsection{Computational costs}

\begin{figure}
    \centering
    \includegraphics[width=0.9\linewidth]{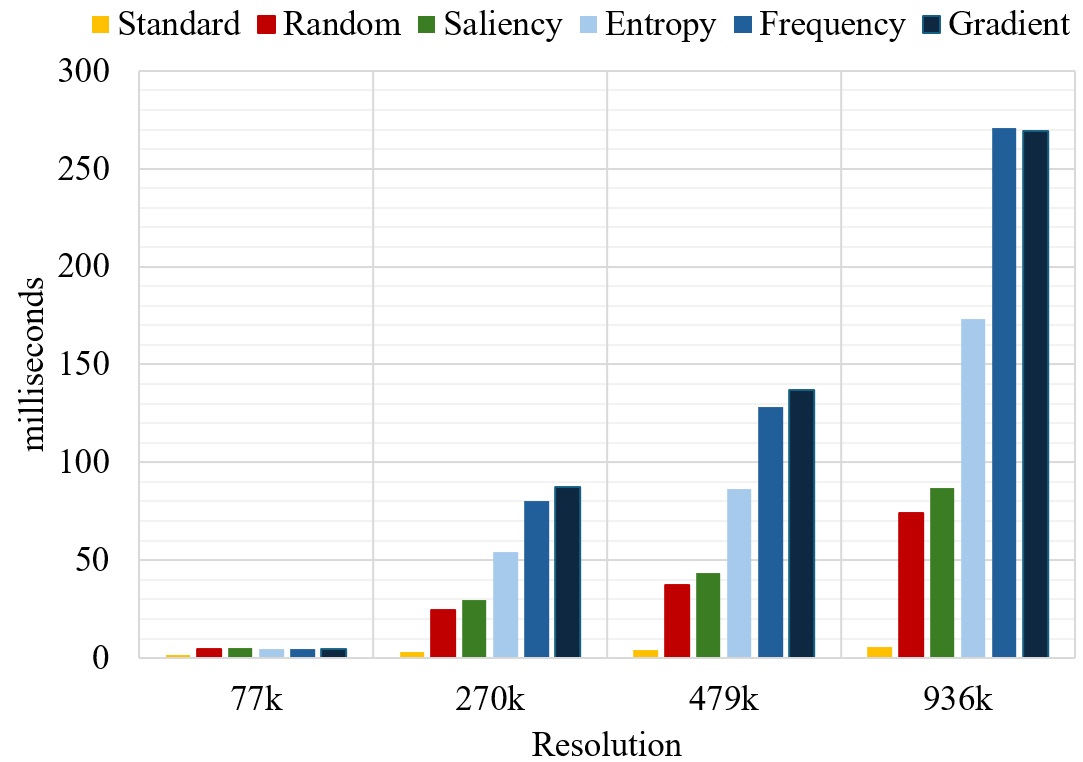}
    \caption{Comparison of \textit{Charm} runtime using different patch selection strategies with the standard approach (Appendix 9). The x-axis shows the total number of pixels.}
    \label{fig:patch_cost}
\end{figure}
Figure \ref{fig:patch_cost} compares the runtime of \textit{Charm} using different patch selection strategies with the standard approach on a 12th Gen Intel(R) Core(TM) i9-12900K CPU. Table \ref{tab:tokenization} demonstrates that multiscale inputs can negatively impact performance for very small images. In this case, we use a single-scale tokenization approach (Section \ref{3.2.1}).
Figure \ref{fig:patch_cost} illustrates the increase in the computational cost of \textit{Charm} with larger image sizes.
To reduce the overhead of saliency mask generation (16 seconds per image), we employ pre-computed masks. The frequency-based method provides a cost-effective alternative to on-the-fly saliency detection, identifying regions comparable to or even superior to those selected by SAM-HQ. Random patch selection emerges as the most effective strategy, offering the best performance with the lowest computational cost.

% \begin{figure}
% \centering
% \includegraphics[width=0.8\linewidth]{figures/model_costs_dino.jpg}
% \caption{Dinov2-small inference cost for processing an image of size 640 x 640. The x-axis and y-axis show the GPU memory in megabytes (MB) and giga multiply accumulation (GMACs), respectively. The disk size represents the inference time. The "Standard" refers to the traditional method of downscaling and cropping images to 224 x 224 (Appendix 8). The "Original Size" indicates the cost of processing the image without any resizing. Exact values can be found in Table 15 in the Appendix.}
% \label{fig:model_cost}
% \end{figure}

The computational cost of ViTs is influenced by the number of input tokens. The standard method involves downscaling and cropping images to 224 x 224, resulting in a very low number of tokens and lower costs. However, processing images at their original size can significantly increase computational costs.
Our approach not only improves performance but also significantly reduces the computational costs of processing images in the original resolution (Table \ref{tab:model_costs}). We observe a decrease of over \textbf{84\%} in GMACs, \textbf{77\%} in runtime, \textbf{83\%} in GPU memory consumption, and \textbf{75\%} in the number of tokens for Dinov2-small when processing one 640 x 640 image using a 2-scale \textit{Charm}. Even the 3-scale version, which is more costly, has significantly lower costs compared to processing images at their original size. 
While \textit{Charm} may incur slightly higher computational costs compared to the standard approach of downscaling and cropping, the substantial performance gains make it a worthwhile trade-off.

\subsubsection{Downscaling factor ($f$)}
Our method initially tokenizes images with an initial patch size ($p'$). Subsequently, selected patches are further tokenized using a smaller patch size ($p$), while the remaining patches are downscaled to $p$. We experiment with different initial patch sizes $p'$=(28, 42, 56), corresponding to downscaling factors ($f$) of 0.5, 0.66, and 0.75. The best performance is achieved with a factor of 0.5 (Appendix 19). 
A larger factor negatively impacts performance, as it leads to more significant downscaling of unselected patches, resulting in information loss for IAA.
\section{Conclusion}
Processing images in their original resolution can be computationally expensive, requiring significant memory and processing power. Also, it is challenging to process images of varying aspect ratios and resolutions due to batch processing constraints.
\textit{Charm} effectively balances computational cost with the preservation of composition, high-resolution details, aspect ratio, and multi-scale information, which are all crucial for achieving optimal performance in IAA. By introducing variability into the input, our method enhances the generalizability of ViTs in IAA. \textit{Charm} is compatible with any ViT architecture (except Swin \cite{liu2021swin}, see Appendix 20) and does not require additional pre-training of ViTs. 
Our method consistently outperforms existing approaches that address similar research questions, achieving promising results across diverse IAA and IQA datasets.

While our approach effectively preserves critical aesthetic-related factors, there's a potential for information loss due to selective downscaling. To further improve performance, future work could explore more sophisticated patch selection strategies, incorporate other modalities, and integrate our method with existing IAA-specific networks.

\textbf{Acknowledgements:} Funded by the European Union (ERC AdG, GRAPPA,
101053925, awarded to Johan Wagemans) and the Research Foundation-Flanders (FWO, 1159925N, awarded to Fatemeh Behrad).

{
    \small
    \bibliographystyle{ieeenat_fullname}
    \bibliography{main}
}
\clearpage
\setcounter{page}{1}
\maketitlesupplementary

% \section{Rationale}
% \label{sec:rationale}
% % 
% Having the supplementary compiled together with the main paper means that:
% % 
% \begin{itemize}
% \item The supplementary can back-reference sections of the main paper, for example, we can refer to \cref{sec:intro};
% \item The main paper can forward reference sub-sections within the supplementary explicitly (e.g. referring to a particular experiment); 
% \item When submitted to arXiv, the supplementary will already included at the end of the paper.
% \end{itemize}
% 
% To split the supplementary pages from the main paper, you can use \href{https://support.apple.com/en-ca/guide/preview/prvw11793/mac#:~:text=Delete%20a%20page%20from%20a,or%20choose%20Edit%20%3E%20Delete).}{Preview (on macOS)}, \href{https://www.adobe.com/acrobat/how-to/delete-pages-from-pdf.html#:~:text=Choose%20%E2%80%9CTools%E2%80%9D%20%3E%20%E2%80%9COrganize,or%20pages%20from%20the%20file.}{Adobe Acrobat} (on all OSs), as well as \href{https://superuser.com/questions/517986/is-it-possible-to-delete-some-pages-of-a-pdf-document}{command line tools}.

\setcounter{section}{5}

\section{Distinct advantages of \textit{Charm} over existing approaches}
\textit{Charm} is not the first approach to focus on processing images at their original resolution. A prominent method in this area is AnyRes \cite{chai2022anyresolutiontraininghighresolutionimage}. To understand the added value of Charm and why it offers a superior approach, we analyzed the Hugging Face implementation of AnyRes (LlavaNextImageProcessor).
While Charm can process images of any size, AnyRes resizes and \textbf{pads} images to a resolution that is 
a multiple of the patch embedding module's input. Then, the images are divided into smaller sub-images, which, along with a downscaled version of the original image, are \textbf{independently} encoded by patch embedding modules. This prevents the model from capturing relationships between smaller sub-images. 
Additionally, AnyRes does not account for cross-scale relationships and treats tokens from different scales equally. In contrast, Charm leverages position and scale embeddings to effectively capture image composition and cross-scale relationships. For batch processing in AnyRes, \textbf{additional padding} is required as images produce different numbers of sub-images.
AnyRes achieved PLCC/SRCC/ACC scores of 0.637/0.619/0.697 on the AADB dataset, which are lower than the standard Dinov2-small tokenizer (0.695/0.682/0.754, respectively).
This is likely due to \textbf{excessive} padding at various stages, which significantly impacts IAA (see Table 2 in the paper).

Closet to Charm are mixed-resolution \cite{ronen2023vision} and mixed-scale tokenization \cite{havtorn2023msvit}. 
To understand the difference between charm and these approaches, consider two versions of a 1024×1024 image: one sharp and the other unsharp. This difference clearly affects their aesthetic scores,
requiring distinct representations for the network to differentiate them. Refs. \cite{ronen2023vision}, \cite{havtorn2023msvit} in the paper first downscale both images to a small fixed size, then retain resolution in some regions while further downsizing others. This produces identical representations for both versions, sufficient for classification but inadequate for aesthetics. In contrast, Charm incorporates high-frequency information from the original image that gets lost with downsampling. Furthermore, while other methods rely on patches from 2 fixed resolutions, Charm is more flexible, learning from patches across varying resolutions. 
Table 3 in the paper shows Charm significantly outperforms these methods (identified as MS).

\section{3-scale Charm}
\label{appendix: 3-scale_charm}
Charm tokenization prepares a sequence of image patches, each with a size of $p\times p \times c$, where $p$ is the patch size of ViT's patch encoding module and $c$ is the number of channels. For low-resolution images, we directly tokenize the image using the patch size of $p$ (as mentioned in Section 3.2.1).

For high-resolution images, we employ a multiscale approach. 
When dealing with 3-scales, we first define our patch sizes as follows:
\begin{equation}
patch\textunderscore sizes = \left\{ \alpha p, \beta p, \gamma p \right\} 
\end{equation}
where $ \gamma > \beta > \alpha, \left\{ \gamma, \beta, \alpha \right\}  \in \mathbb{N}$.
The maximum downscaling is $f = \alpha / \gamma$. 
% a set of downscaling factors based on the desired number of scales ($ns$) and the maximum downscaling factor ($f$) using Equation \ref{eq:factors}. 

% \begin{equation}
% \label{eq:factors}
% factors = \bigcup_{s \in \left\{0,...,ns-1 \right\}} \left\{ s \times \left (  \frac{1-f}{ns-1} \right ) + f   \right\} 
% \end{equation}

% For instance, with 3 scales ($ns = 3$) and $f=0.5$, factors would be 0.5, 0.75, 1. These factors determine the patch sizes at each scale (Equation \ref{eq:patch_size}).

% \begin{equation}
% \label{eq:patch_size}
%     patch\textunderscore sizes = \bigcup_{f' \in factors} \left\{ f' \times 2^{ns-1} \times p \right\}  
% \end{equation}
% For example, with a $p=14$ (as in dinov2-small) and a scaling factor of $f=0.5$, the three scales would have patch sizes of 28, 42, and 56.
The image is initially tokenized using the largest patch size ($\gamma p$). A subset of patches is then selected and further tokenized using the base patch size ($p$). 
We then select another subset of patches from unselected regions for the second scale. These selected regions are downscaled to the intermediate patch size ($\beta p$) and then tokenized using $p$.
Finally, the remaining regions are downscaled to the smallest patch size ($\alpha p$) and then further tokenized using $p$.  

As discussed in Section 4.4.5, a scaling factor of $f = 0.5$ yields optimal results. Consequently, we employ patch sizes of {2p, 3p, 4p} for our 3-scale version. Table \ref{tab:scale} indicates that the 2-scale \textit{Charm} tokenization yields the best performance on AVA and AADB datasets.

\begin{table}
\renewcommand\thetable{6}
\small
    \centering
    \begin{tabular}{cccccc}
    \hline
        Dataset & Charm & Scale & PLCC & SRCC & ACC\\
        \hline
         \multirow {3}{*}{AVA}
         & - & 1 & 0.710 & 0.706 & 0.802 \\
         & \checkmark & 2 & \textbf{0.779} &\textbf{0.779} & \textbf{0.826} \\
         & \checkmark & 3 & 0.759 & 0.757 & 0.817 \\ 
         \hline
         \multirow {3}{*}{AADB}
         & -  & 1 & 0.695 & 0.682 & 0.754 \\ 
         &\checkmark &  2 & \textbf{0.767} & \textbf{0.754} & 0.767 \\
         & \checkmark & 3  & 0.753 & 0.745 & \textbf{0.775} \\ 
         \hline
    \end{tabular}
    \caption{Performance of Dinov2-small on AVA and AADB datasets across different scales.}
    \label{tab:scale}
\end{table}

\section{Patch selection strategies}
\label{appendix: patch_selection}
To identify important areas for patch selection, we explore various strategies. The goal is to preserve visually interesting regions and sharp details in high resolution.

We initially consider using saliency maps generated by the SAM-HQ model \cite{ke2024segment} with different prompts. These prompts include ``figure ground reversal,'' ``figure ground separation,'' ``figure ground segmentation,'' ``camo object,'' ``salient object,'' and ``hidden object.'' All of these prompts represent the figure-ground organization, where humans simplify a scene into the main object (the figure) and everything else (the background) \cite{peterson2014low, wagemans2012century}. The results show that the 'salient object' is the most effective prompt. After creating saliency maps, we randomly select patches within the salient area in each epoch.

\begin{figure}
\renewcommand\thefigure{7}
    \centering
    \includegraphics[width=1\linewidth]{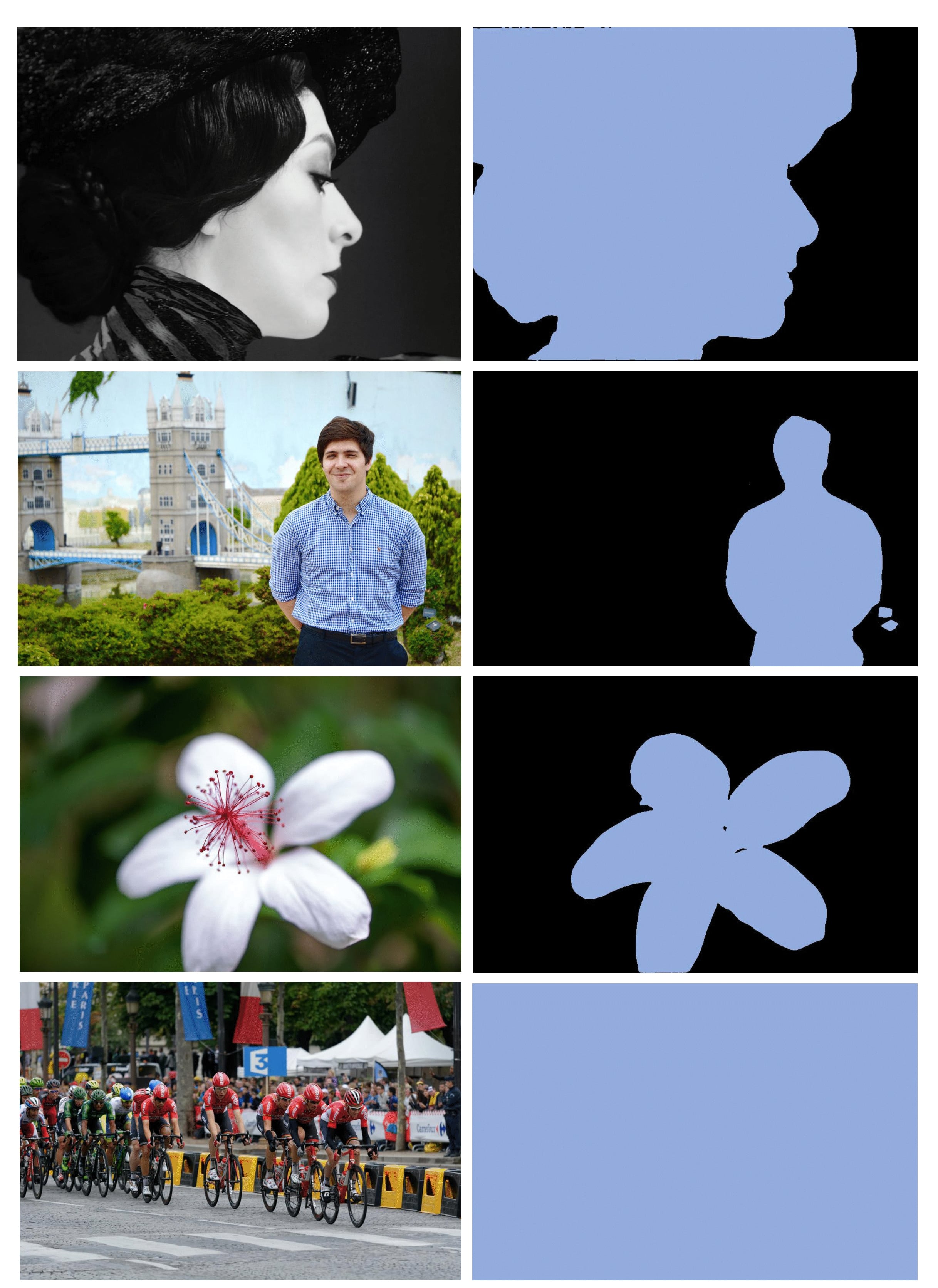}
    \caption{Examples of saliency maps generated by SAM-HQ with the prompt of ``salient object''. The final row shows a failure case where the Sam-HQ model did not detect any salient object.}
    \label{fig:sam-hq}
\end{figure}
Figure \ref{fig:sam-hq} shows examples of the salient objects identified by the model. In rare cases, the model fails to identify any salient objects. In such instances, we simply consider the entire image as salient, ensuring that patches are selected from all areas. The rate of model failures is negligible, \textbf{less than 0.14\%} in the AVA dataset.

\begin{figure*}
\renewcommand\thefigure{8}
    \centering
    \includegraphics[width=1\linewidth]{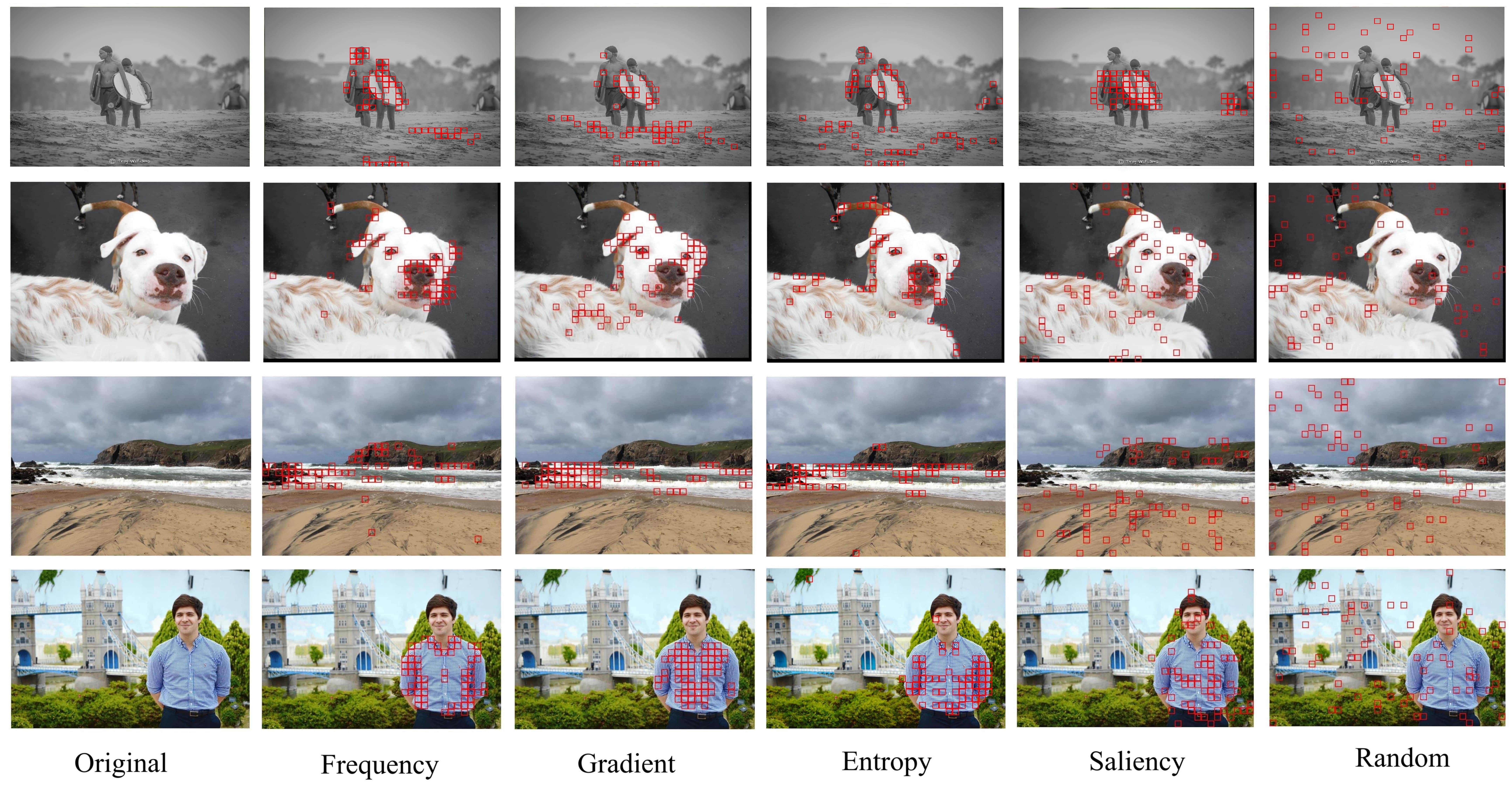}
    \caption{Visualization of patch selection strategies. Red squares highlight areas selected by different methods. Frequency, gradient, and entropy approaches are fully deterministic in these examples.}
    \label{fig:patch_selection}
\end{figure*}
\begin{figure*}
\renewcommand\thefigure{9}
    \centering
    \includegraphics[width=0.75\linewidth]{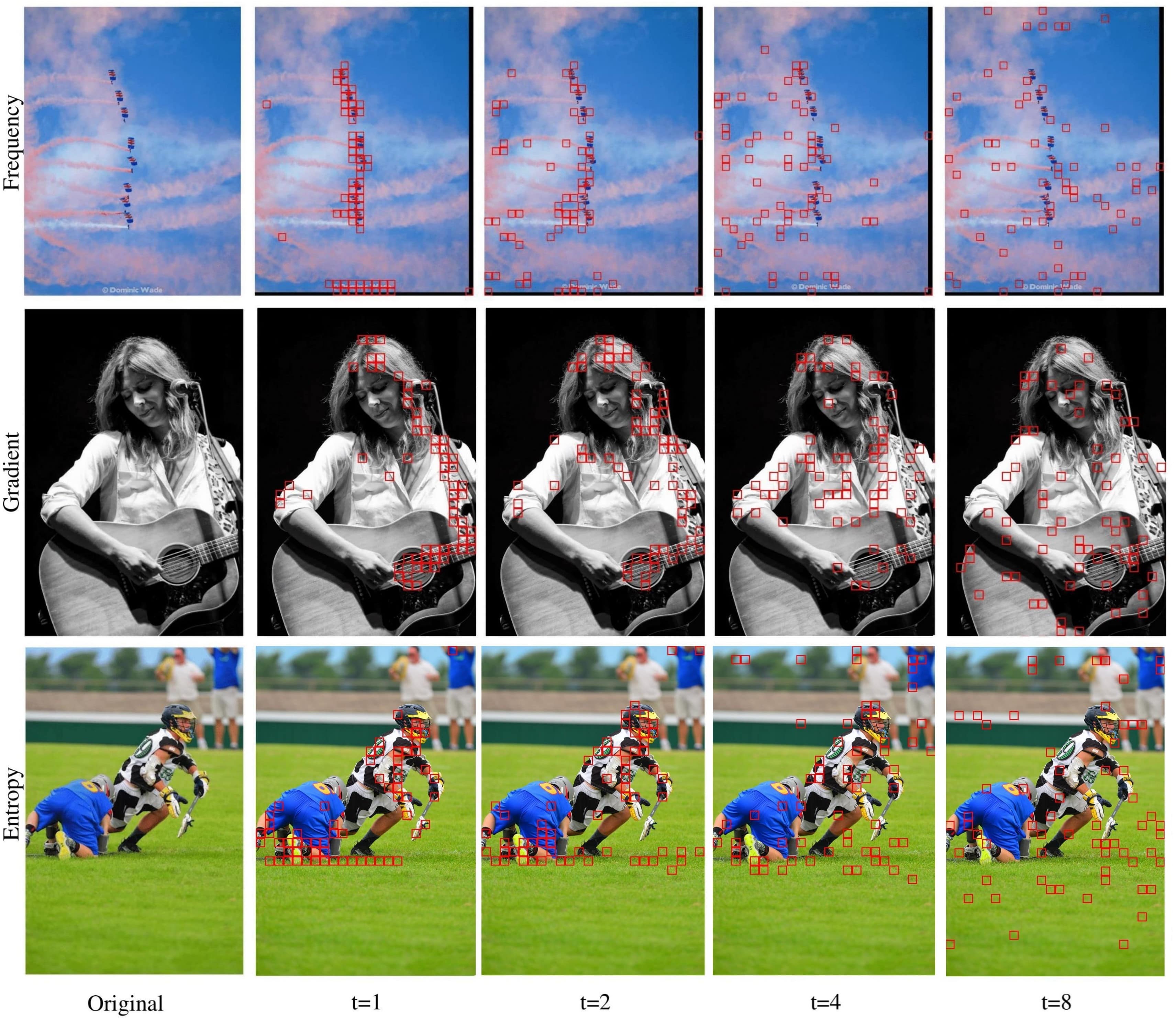}
    \caption{The impact of different thresholds ($t$) on the frequency, gradient, and entropy-based patch selection. Increasing the value of $t$ introduces more diversity to the selected patches.
    }
    \label{fig:threshold}
\end{figure*}

We also consider other methods, such as frequency, gradient, and entropy, to identify important regions. These metrics are calculated using the Fast Fourier transform, Sobel filter, and Shannon entropy, respectively.
We start with a fully deterministic approach, selecting only patches with the highest frequency, entropy, or gradient. We then investigate whether introducing a degree of randomness can improve performance. To balance determinism and randomness, we introduce a threshold parameter ($t$). To select patches, we first rank them based on their frequency, entropy, or gradient values. We then select a larger number of patches ($t$ times the desired number) with the highest scores. Finally, we randomly select the required number of patches from this pool in each epoch. As shown in Figure \ref{fig:threshold}, smaller values of $t$ result in lower variability and higher correlation between selected patches.

\begin{figure*}
\renewcommand\thefigure{10}
    \centering
    \includegraphics[width=0.8\linewidth]{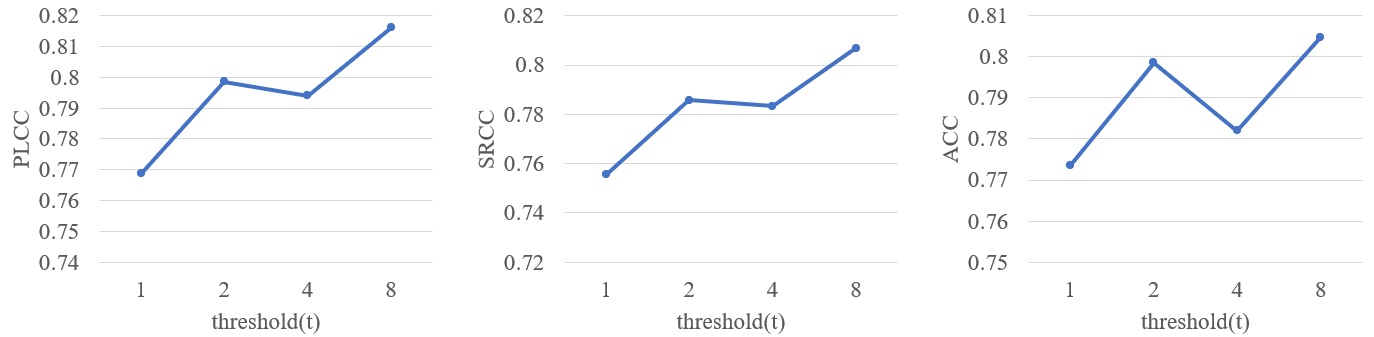}
    \caption{The performance of Dinov2-small + \textit{Charm }on the AADB dataset using gradient-based patch selection with different thresholds ($t$). The results are averaged over 5 runs on the validation set. Generally, increasing the threshold ($t$) leads to improved performance.  
    }
    \label{fig:threshold_results}
\end{figure*}

We fine-tune Dinov2-small on the AADB dataset using different thresholds. Figure \ref{fig:threshold_results} indicates that increasing the threshold t generally improves validation performance. However, we use $t=2$ to balance the inclusion of high-frequency, entropy, and gradient-based patches while maintaining diversity in the selected regions.

We also consider random patch selection. Table \ref{tab:random_fixed} demonstrates that randomly selecting patches in the first epoch and keeping them fixed throughout training yields better results than the standard Dinov2-small (around 1 \% improvement). However, by randomly selecting patches in each epoch, we can achieve further performance improvements, surpassing other approaches (Table \ref{tab:patch_selection_extended}). Randomly sampling patches in each epoch exposes the ViT model to different regions of the image in high resolution. This diversity in training data helps prevent overfitting. 
Red squares in Figure \ref{fig:patch_selection} demonstrate the selected patches using our patch selection strategies.

\begin{table}
\small
\renewcommand\thetable{7}
    \centering
    \begin{tabular}{lccc}
    \hline
        Tokenization & PLCC & SRCC & ACC \\
    \hline
        Standard  & 0.695 & 0.682 & 0.754 \\
        Charm + Random (first epoch) & 0.705 & 0.693 & 0.756 \\ 
        Charm + Random (each epoch) & \textbf{0.767} & \textbf{0.754} &\textbf{ 0.767} \\
    \hline
    \end{tabular}
    \caption{
    Performance of Dinov2-small on the AADB dataset using different tokenization approaches. \textit{Charm} with random patch selection in each epoch achieves the best performance.}
    \label{tab:random_fixed}
\end{table}

\begin{table*}
\renewcommand\thetable{8}
    \centering
    \begin{tabular}{llcccccc}
    \hline
        \multirow{2}{*}{Patch selection} & \multirow{2}{*}{Charm} & \multicolumn{2}{c}{PLCC}& \multicolumn{2}{c}{SRCC} & \multicolumn{2}{c}{ACC}\\ \cline{3-8} & & train & test & train & test & train & test \\
        \hline
        % \multicolumn{7}{c}{\cellcolor{gray!25}AADB} \\
        % \hline
        Random & \checkmark & 0.800 & \textbf{0.767} & 0.794 & \textbf{0.754} & 0.801 & \textbf{0.767} \\
        Frequency & \checkmark & 0.831 &\underline{0.756} & 0.824 & \underline{0.747} & 0.815 & 0.761 \\
        Entropy & \checkmark & 0.891 & 0.726 & 0.887 & 0.714 & 0.823 & 0.761 \\
        Gradient & \checkmark & 0.838 & 0.734 & 0.845 & 0.721 & 0.825 & \underline{0.766} \\
        Saliency & \checkmark & 0.870 & 0.751 & 0.862 & 0.738 & 0.835 & 0.756 \\
        Standard & - & 0.768 & 0.695 & 0.753 & 0.682 & 0.773 & 0.754 \\
        \hline
        % \multicolumn{7}{c}{\cellcolor{gray!25}AVA} \\
        % \hline
        % Random & 0.801 & \textbf{0.779} & 0.791 & \textbf{0.777} & 0.843 & \textbf{0.826} \\
        % Frequency & 0.784 & 0.774 & 0.774 & 0.773 & 0.836 & 0.823 \\
        % Entropy & 0.827 & \underline{0.776} & 0.821 & \underline{0.775} & 0.856 & \underline{0.826} \\
        % Standard & 0.726 & 0.710 & 0.714 & 0.706 & 0.812 & 0.802 \\ \hline
    \end{tabular}
    \caption{The performance of Dinov2-small on the AADB dataset using different patch selection approaches. \textbf{Bold} and \underline{underlined} numbers represent the best and second-best results. This table is the extended version of Table 4. We use $t=2$ to ensure a diverse set of patches with high frequency, entropy, and gradient.}
    \label{tab:patch_selection_extended}
\end{table*}

% \begin{figure}
%     \centering
%     \includegraphics[width=1\linewidth]{figures/PositionEmbedding.png}
%     \caption{For high-resolution images, two distinct position embeddings are generated, one for low-resolution patches (a) and another for high-resolution patches (b). These embeddings are then aggregated to create a final position embedding (c) incorporating information from both scales.}
%     \label{fig:position_embedding}
% \end{figure}

\section{Fine-tuning Dinov2-small for IAA}
\label{appendix: original_dino}
\begin{figure}
\renewcommand\thefigure{11}
    \centering
    \includegraphics[width=0.65\linewidth]{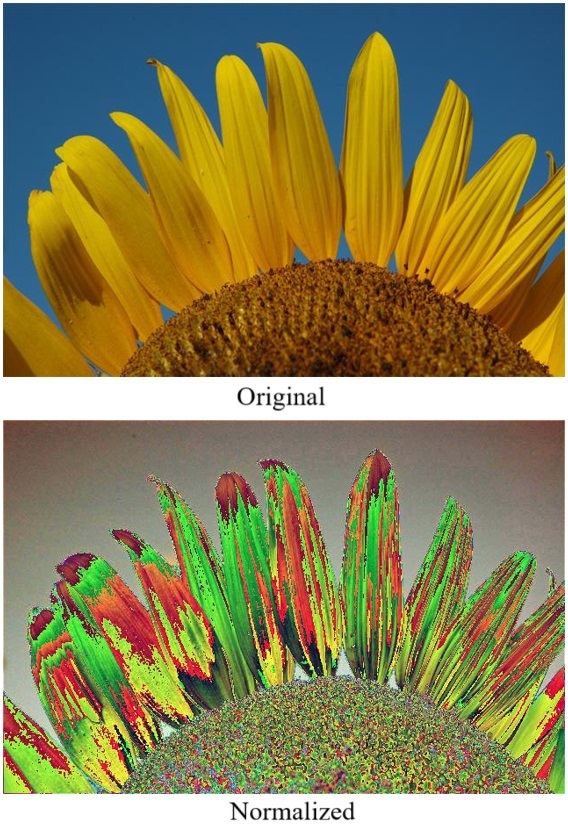}
    \caption{The effect of normalization on an image. Normalization can introduce visual artifacts and distort the aesthetic quality of an image.}
    \label{fig:normalization}
\end{figure}

The original Dinov2 code downscales images to the shortest edge of 256 and applies center cropping to create fixed input sizes ($224 \times 224$). Additionally, images are normalized, a common practice in deep learning to help the network learn faster and better. However, we observed that normalization can negatively impact IAA performance (Table \ref{tab:normalization}) due to significant changes in the images (Figure \ref{fig:normalization}). As a result, we remove normalization. Also, after downscaling images to the shortest edge of 256, we use random cropping instead of center cropping. As shown in Table \ref{tab:dinov2}, our approach outperforms the original Dinov2-small setting for IAA. Throughout this paper, the 'Standard approach' refers to our settings.

\begin{table}
\renewcommand\thetable{9}
    \centering
    \begin{tabular}{lccc}
    \hline
        Normalization & PLCC & SRCC & ACC \\
    \hline
        True & 0.474 & 0.446 & 0.664\\
        False & \textbf{0.488} & \textbf{0.458} &\textbf{ 0.794} \\
    \hline
    \end{tabular}
    \caption{The effect of image normalization on the performance of Dinov2-small + \textit{Charm} on the TAD66k dataset. Image normalization negatively affects the IAA.}
    \label{tab:normalization}
\end{table}
\begin{table}
\renewcommand\thetable{10}
    \centering
    \begin{tabular}{lccc}
        \hline
        Implementation & PLCC & SRCC & ACC \\
        \hline
        Standard & 0.697 & 0.694 & 0.799 \\
        Our settings &\textbf{0.710} & \textbf{0.706} & \textbf{0.802} \\ \hline
    \end{tabular}
    \caption{Performance of Dinov2-small on the AVA dataset with two different data preprocessing approaches.}
    \label{tab:dinov2}
\end{table}

\section{Data augmentation methods}
\label{appendix:data_augmentation}
\begin{figure}
\renewcommand\thefigure{12}
    \centering
    \includegraphics[width=1\linewidth]{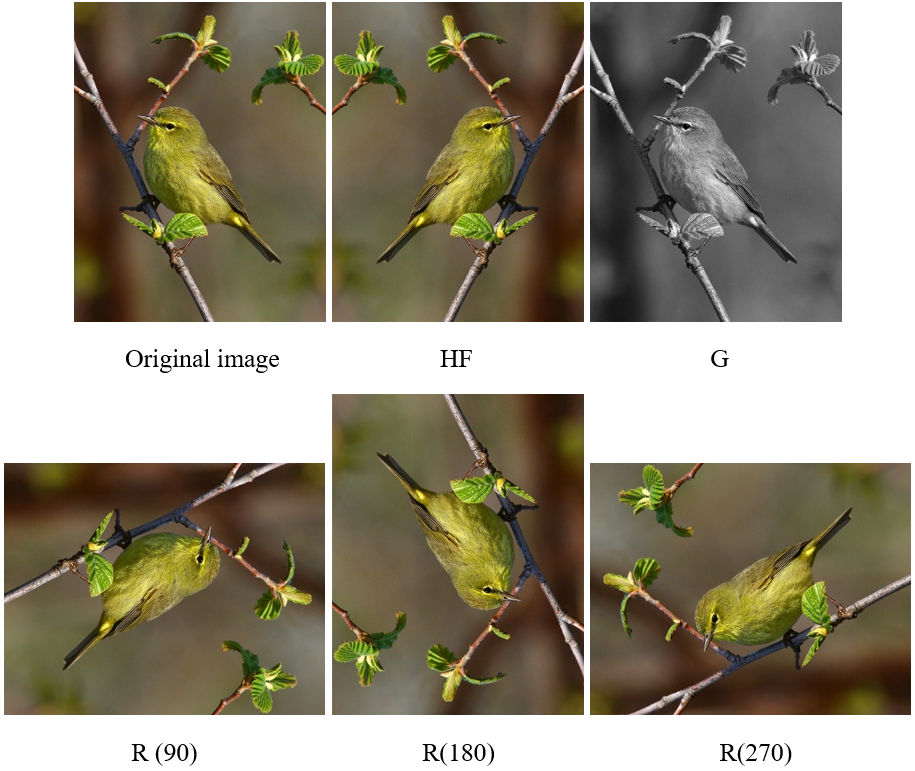}
    \caption{Data augmentation methods that preserve the composition of elements in the image. HF, G, and R represent horizontal flipping, grayscale augmentation, and random rotation.}
    \label{fig:augmentation}
\end{figure}
Some argue that existing global data augmentation techniques, which alter the entire image, can potentially change the aesthetic labels and should thus be avoided in IAA \cite{strafforello2024backflip}. However, ViTs are prone to overfitting and often require large datasets for fine-tuning, which can be challenging in IAA.
Among existing global data augmentation methods, horizontal flipping, random rotation (at angles of 90, 180, or 270 degrees), and grayscale augmentation preserve the composition of elements. We apply these methods to evaluate their impact on model performance. Figure \ref{fig:augmentation} shows examples of these augmentations, which are applied with a probability of 50\%.

While these augmentations may affect human aesthetic judgment, they consistently improve the generalizability of ViTs (as shown in Table \ref{tab:augmentation}). Grayscale augmentation slightly decreases performance, highlighting the importance of color in IAA. Therefore, we only use random horizontal flipping and random rotation in our experiments.

\begin{table*}
\renewcommand\thetable{11}
    \centering
    \begin{tabular}{lcccccc}
    \hline
        \multirow{2}{*}{Augmentation} & \multicolumn{2}{c}{PLCC}& \multicolumn{2}{c}{SRCC} & \multicolumn{2}{c}{ACC}\\ \cline{2-7} & train & test & train & test & train & test \\
        \hline
        No augment & 0.833 & 0.768 & 0.829 & 0.766 & 0.859 & 0.822 \\
        HF & 0.818 & 0.776 & 0.812 & \underline{0.775} & 0.851 & \underline{0.824} \\
        HF + G & 0.823 & 0.774 & 0.817 & 0.772 & 0.855 & \underline{0.824} \\
        HF + R & 0.801 & \textbf{0.779} & 0.791 & \textbf{0.777} & 0.843 & \textbf{0.826} \\
        HF + R + G & 0.783 & \underline{0.778} & 0.772 & \textbf{0.777} & 0.834 & \underline{0.824}\\
        \hline
    \end{tabular}
    \caption{The performance of Dinov2-small + \textit{Charm} on the AVA dataset using different data augmentation methods. HF, G, and R represent horizontal flipping, grayscale augmentation, and random rotation, respectively. \textbf{Bold} and \underline{underlined} numbers represent the best and the second-best results.}
    \label{tab:augmentation}
\end{table*}

\section{Comparing dataset resolutions}
\label{appendix: resolution}
\begin{figure}
\renewcommand\thefigure{13}
    \centering
    \includegraphics[width=1\linewidth]{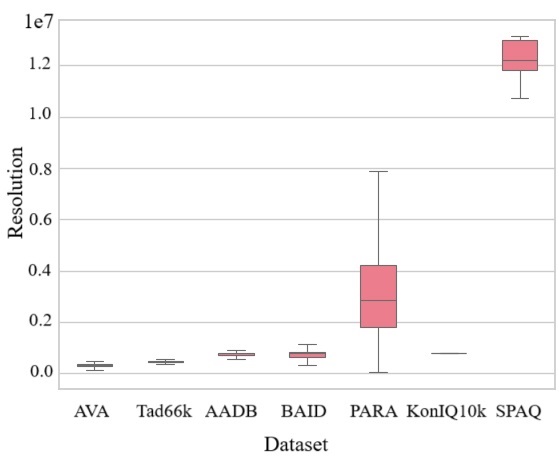}
    \caption{Distribution of image resolutions across datasets.}
    \label{fig:resolution}
\end{figure}
We compare the image resolutions within both IAA and IQA datasets. Resolution represents the number of pixels in the image and is calculated by multiplying the width and height of images.
As shown in Figure \ref{fig:resolution}, PARA and SPAQ have the highest variety and resolution compared to the others. 
Among the other datasets, AVA has the lowest resolution, and all datasets exhibit a normal distribution of image resolutions. As all images in KonIQ10k have a fixed resolution of 1024 × 768 pixels, it appears as a single line in the box plot. Our analysis shows no significant differences in image resolution between the training and test sets of the datasets.

\section{Importance of scale embedding}
A key focus of this paper is to preserve the compositional relationships between elements. To achieve this, we avoid cropping and use position and scale embeddings to capture the relationships between tokens both across and within different scales. 
Adding scale embedding to Dinov2-small with the Charm tokenizer increases PLCC/SRCC/ACC from 0.748/0.739/0.750 to 
0.767/0.754/0.767 on the AADB dataset. This 2\% improvement surpasses the scale embedding gains reported in Ref. \cite{ke2021musiq}.

\section{The maximum number of patches ($l$)}
\label{appendix: input_length}
We conduct ablation studies with different input lengths ($l$) during training. The optimal value for $l$ should be chosen based on the image resolutions in the dataset.
Setting $l$ larger than the average number of patches (average of $s$ across images) can lead to excessive padding while setting it smaller can result in excessive cropping. Both scenarios can negatively impact performance in IAA. 
For example, in the AADB dataset, the average number of patches after preprocessing using \textit{Charm} is 1090. As shown in Table \ref{tab:input_length}, an input length ($l$) of 1024, which is closest to the average number of tokens, yields the best performance for predicting the aesthetic score.

\begin{table}
\renewcommand\thetable{12}
    \centering
    \begin{tabular}{lccc}
    \hline
        Input length($l$) & PLCC & SRCC & ACC \\
    \hline
        768 & 0.743 & 0.739 & 0.770\\
        1024 & \textbf{0.767} & \textbf{0.754} & 0.767\\
        1500 & 0.736 & 0.727 & \textbf{0.775}\\
    \hline
    \end{tabular}
    \caption{The impact of input length ($l$) on performance of Dinov2-small + \textit{Charm} on the AADB dataset. Selecting the optimal $l$ value is crucial for achieving the best results.}
    \label{tab:input_length}
\end{table}

\section{Charm's influence on ViT-based models}
In this section, we demonstrate that incorporating Charm enhances the performance of various ViT models. We evaluated its effectiveness on different backbones, including ViT-Small, Dinov2-Small, and Dinov2-Large. As shown in Table \ref{tab:backbones}, Charm significantly improves performance on the AVA dataset across all tested models. These results indicate that Charm is not dependent on a specific architecture and performs consistently well across both smaller and larger models.

\begin{table}
\renewcommand\thetable{13}
\small
    \centering
    \begin{tabular}{m{4em}m{1cm}m{4em}m{4em}m{4em}} 
        \hline
        Model & Charm & PLCC & SRCC & ACC \\
        \hline
        \multirow {2}{*}{\makecell{Dinov2\\-small}} & - & 0.710 & 0.706 & 0.802 \\ \cline{2-5}
        & \checkmark & 0.779 \color{NavyBlue}{($\uparrow6.9\%$)}
        & 0.777 \color{NavyBlue}{($\uparrow7.1\%$)} & 0.826 \color{NavyBlue}{($\uparrow2.4\%$)} \\
        
        \hline
        \multirow {2}{*}{ViT-small} & - & 0.687 & 0.679 & 0.794 \\ \cline{2-5}
        & \checkmark & 0.762 \color{NavyBlue}{($\uparrow7.5\%$)} & 0.760 \color{NavyBlue}{($\uparrow8.1\%$)} & 0.827 \color{NavyBlue}{($\uparrow3.3\%$)}\\

        \hline 
        \multirow {2}{*}{\makecell{Dinov2\\-large}} & - & 0.734 & 0.732 & 0.808 \\ \cline{2-5}
        & \checkmark & 0.783 \color{NavyBlue}{($\uparrow4.9\%$)} & 0.781 \color{NavyBlue}{($\uparrow4.9\%$)} & 0.828 \color{NavyBlue}{($\uparrow2\%$)}\\
        
        \hline
    \end{tabular}
    \caption{Performance improvement across different models on the AVA dataset by replacing their standard tokenization with \textit{Charm}. All experiments using \textit{Charm} employ a random patch selection strategy.}
    \label{tab:backbones}
\end{table}

\section{Super high-resolution images}
\label{appendix: super resolution}
As shown in Figure 6 and Table 5 in the paper, increasing the resolution can increase the computational costs of our method. This is especially challenging when dealing with extremely large images (e.g., 3k by 4k pixels).
However, our experiments on the PARA dataset demonstrate that there is a threshold for performance improvement due to preserving high-resolution information. Beyond this threshold, further performance gains are limited. By downscaling images to a maximum edge of 1024, we can significantly reduce computational costs without sacrificing much performance. Table \ref{tab:super_large} shows that the performance difference between downscaling to the maximum edge of 1024 and 1500 is less than 0.3\% in PLCC and SRCC and only 0.8\% in ACC.
This suggests that processing images at excessively high resolutions may not provide significant benefits, especially considering the increased computational costs. 

\begin{table}
\renewcommand\thetable{14}
    \centering
    \begin{tabular}{lccc}
    \hline
        Approach & PLCC & SRCC & ACC \\
        \hline
        Standard & 0.904 & 0.855 & 0.863 \\
        Maximum edge = 1024 & 0.938 & 0.905 & 0.892 \\
        Maximum edge = 1500 & \textbf{0.940} & \textbf{0.908} & \textbf{0.900 }\\
        \hline
    \end{tabular}
    \caption{Performance of Dinov2-small on PARA dataset with different resolution. ``Standard'' refers to the approach explained in Appendix \ref{appendix: original_dino}. The others use Charm. Processing images in high resolution positively affects the performance of the IAA model.}
    \label{tab:super_large}
\end{table}

\section{Dinov2 + Muller / Padding}
\label{appendix:muller}
\begin{figure}
\renewcommand\thefigure{14}
    \centering
    \includegraphics[width=1\linewidth]{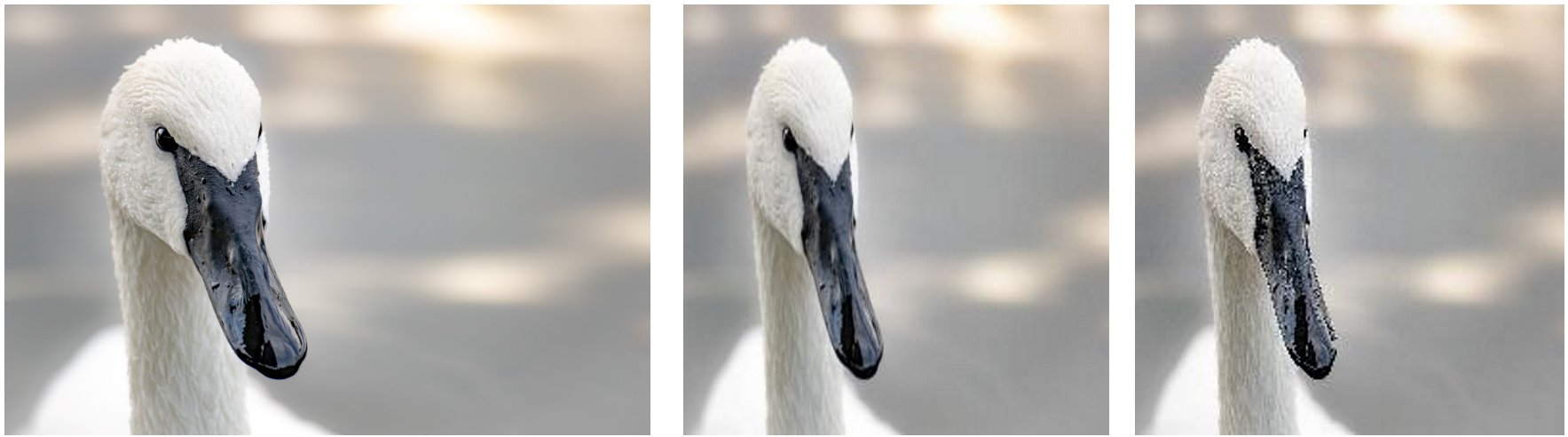}
    \caption{From left to right: the original image, downscaled image by bilinear interpolation, and downscaled image by Muller. While effective in image classification, the Muller method can introduce distortions that negatively impact aesthetics.}
    \label{fig:muller}
\end{figure}

Muller \cite{tu2023muller} is a learnable resizer that aims to boost details in certain frequency subbands during downscaling. While effective in image classification, Muller can introduce distortions (Figure \ref{fig:muller}) that negatively affect aesthetics (Table 2).

\begin{figure}
\renewcommand\thefigure{15}
    \centering
    \includegraphics[width=1\linewidth]{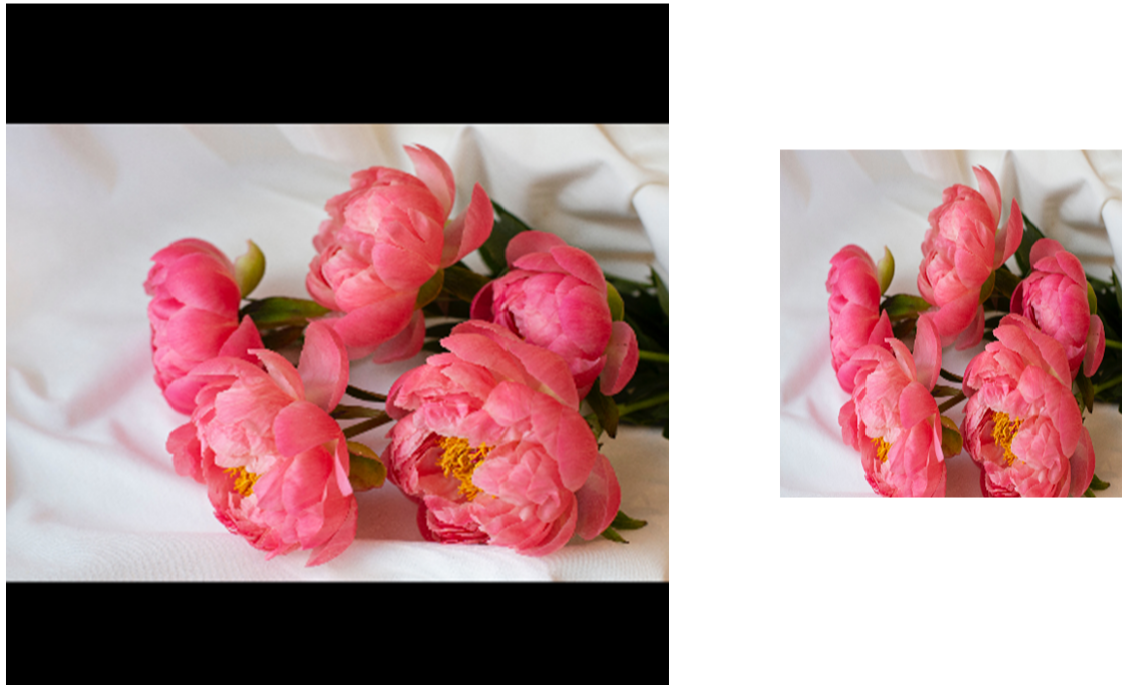}
    \caption{Comparison of Dinov2-small standard input (Section \ref{appendix: original_dino}) and padded input (Section \ref{appendix:muller}). While padding preserves the aspect ratio, it can negatively impact the performance of IAA models.}
    \label{fig:padding}
\end{figure}
A straightforward approach to preserving the aspect ratio is to add padding to images. For this approach, we resize images to a maximum edge of 512 and add padding (Figure \ref{fig:padding}). 
While this approach preserves the aspect ratio and maintains higher-resolution images compared to the standard method, it yields worse results (Table 2), highlighting the significant negative impact of padding in IAA.
Moreover, padding not only fails to add valuable information but also significantly increases the training costs of the standard DinoV2-small model.

\section{Detailed comparison of our approach with state of the art in IAA and IQA}
\label{appendix: extended_analysis}

\begin{table*}
\renewcommand\thetable{15}
\small
    \centering
    \begin{tabular}{m{7em}cccccccccccc}
    \hline
         \multirow{2}{*}{Algorithm} & \multirow{2}{*}{MM} & \multicolumn{2}{c}{AVA} & \multicolumn{2}{c}{AADB} & \multicolumn{2}{c}{TAD66k} & \multicolumn{2}{c}{PARA} & \multicolumn{2}{c}{BAID} & \multirow{2}{*}{\# params} \\ 
         \cline{3-12} & & PLCC & SRCC & PLCC & SRCC & PLCC & SRCC & PLCC & SRCC& PLCC & SRCC & \\
         \hline
         \cite{10054147} & \checkmark & 0.736&0.725&0.763&0.761&-&-&0.940&0.911&-&-&48.84 M \\
         \cite{10502339}& \checkmark &0.753&0.751&\underline{0.770}&0.768&-&-&-&-&-&-&87 M\\
         \cite{10433197}& \checkmark & 0.754&0.752&-&-&-&-&0.928&0.895&-&-&76.7 M\\
         \cite{sheng2023aesclip}& \checkmark & 0.779 & 0.771 & - & \textbf{0.79} & -  & - & \textbf{0.951} & \textbf{0.926} & - & - & 149.6 M \\
         \cite{li2024towards}& \checkmark &0.785& 0.776&	- & \underline{0.771}  & - & - & - & - &- & - & 3.149 B \\
         \cite{10115482}& \checkmark& \underline{0.83} & \underline{0.816} & - & - & - & -& - & - & - & - & 158.8 \\
         
         % \cite{zhou2024uniaa}& \checkmark& \textbf{0.838} & \textbf{0.84} & - & - & \textbf{0.553} & 0.521 & - & - & - & - & 7 B \\
         \cite{zhu2024emotion} & \checkmark & \textbf{0.834}&\textbf{0.819}& - & - & - & - & - & - & - & - & 135.5 B \\
         \cline{2-13}
         \cite{shi2024improving}& - & 0.758 & 0.758 & - & - & \textbf{0.553} & \underline{0.530} & - & - & \textbf{0.558} &\textbf{ 0.508} & 56 M \\
         \cite{liuelta}& - & 0.777 & 0.764 & \textbf{0.772} & 0.760 & 0.539 & 0.496 & \underline{0.943} & \underline{0.912} & - & - & 3 B \\
         \cite{he2023eat} & - & 0.814 & 0.803 & - & - & \underline{0.546} & \textbf{0.57} & - & - & - & - & 87 M \\
         \cite{yi2023towards} &-&-&-&-&-&-&-&-&-&\underline{0.473}& \underline{0.467} & \underline{27.3 M} \\
         Dinov2-small + Charm & -&0.779&0.777&0.767&0.754&0.488&0.458&0.940&0.908&0.439&0.368&\textbf{21.53 M} 
         
         % ViT-Base + Charm &-&-&-&-&-&-&-&-&-& \underline{0.515}& \underline{0.467}& 87.4 M 
         \\ \hline
    \end{tabular}
    \caption{Detailed comparison of our approach with existing IAA models. MM represents using multimodal data like text and attributes. Our approach achieves comparable performance to state-of-the-art IAA models while using significantly fewer parameters.}
    \label{tab:detatiled_comparison_sota}
\end{table*}

Table \ref{tab:detatiled_comparison_sota} and \ref{tab:detatiled_comparison_sota_iqa} illustrate the detailed comparison of our approach with state-of-the-art models in IAA and IQA. 
Charm focuses on crucial image-based factors to improve the performance of ViTs, which are the foundation of many state-of-the-art IAA models.
While the use of multimodal models is orthogonal to Charm’s contribution, it is likely that integrating Charm in a multimodal method will lead to improved performance. Unfortunately, none of these methods (refs.  \cite{10502339},\cite{sheng2023aesclip}, \cite{li2024towards}, and \cite{zhu2024emotion}) have code available to test this. Other methods (refs. \cite{10054147}, \cite{10433197}, and \cite{10115482}) are CNN-based.

Additionally, Our approach achieves comparable performance while having considerably fewer parameters (Figure 4). Our approach adds only 1152 parameters to the Dinov2-small model for the scale embedding (1, 2, 384) and mask token (1, 1, 384).
The mask token is a learnable parameter used in the masking process of the input. In batch training, masks are used to identify effective inputs while ignoring padding tokens that may be present in some images.

\begin{table}
\renewcommand\thetable{16}
\small
    \centering
    \begin{tabular}{cccccc}
    \hline
         \multirow{2}{*}{Algorithm} & \multicolumn{2}{c}{SPAQ} & \multicolumn{2}{c}{KonIQ10k} & \multirow{2}{*}{\# params} \\ 
         \cline{2-5} & train & test & train & test & \\
         \hline
         \cite{shin2024blind}& \textbf{0.928}&\textbf{0.923}&\textbf{0.945}&\textbf{0.934}&30.97 M\\
         \cite{chen2024topiq}&0.924&0.921&0.939&0.926&25.6 M\\
         \cite{ke2021musiq}& 0.921&0.917&0.928&0.916&27 M\\
         Ours &0.919&0.915&0.944&0.930&\textbf{21.53 M} \\ \hline
    \end{tabular}
    \caption{Detailed comparison of our approach with existing IQA models. Our approach falls slightly behind state-of-the-art methods in IQA, with a difference of less than 1\% in SPAQ and 0.4\% in KonIQ10k. While other studies have reported the median of 10 runs, we conducted 5 runs and found consistent results, with standard deviations below 0.006 for SPAQ and 0.002 for KonIQ10k.}
    \label{tab:detatiled_comparison_sota_iqa}
\end{table}

\begin{table*}
\renewcommand\thetable{17}
\centering
    \begin{tabular}{lllllll}
        \hline
        Model & Input size& Charm & \#tokens & ms & GMACs & MB \\
        \hline
        \multirow {2}{*}{ViT-small} & 224 x 224 & - &196 & \textbf{5.6} & \textbf{4.58} & \textbf{168.4} \\ \cline{2-7}  &
        \multirow {3}{*}{640 x 640} & - & 1600 & 23.6 & 58.11 & 1352.1 \\ && \checkmark & 2-scale:512 & \underline{7.1}\color{NavyBlue}{($\downarrow69.9\%$)}  & \underline{13.49}\color{NavyBlue}{($\downarrow76.8\%$)}  & \underline{328.3}\color{NavyBlue}{($\downarrow75.7\%$)}  \\ && \checkmark & 3-scale:700 & 9.1\color{NavyBlue}{($\downarrow61.4\%$)} & 19.65\color{NavyBlue}{($\downarrow66.2\%$)}  & 469.3\color{NavyBlue}{($\downarrow65.3\%$)}  \\ 
         \hline
    \end{tabular}
    \caption{ViT-small inference cost breakdown for processing one single image: number of tokens (\#tokens) based on varying input sizes, runtime in milliseconds (ms), Giga multiply accumulation (GMACs), and GPU memory in Megabytes (MB). \textbf{Bold} and \underline{underlined} values highlight the most and second-most computationally efficient configurations.
    Percentages indicate the reduction in computational cost compared to processing the image in its original size.}
    \label{tab:model_costs_extended}
\end{table*}

\begin{table}
\renewcommand\thetable{18}
    \centering
    \begin{tabular}{llccc}
    \hline
        $p'$ & $f$ & PLCC & SRCC & ACC \\
        \hline
        28 & 0.5& \textbf{0.767} & \textbf{0.754} & 0.767 \\
        42 & 0.66 & 0.732 & 0.725 & \textbf{0.774} \\
        56 & 0.75& 0.739 & 0.729 & 0.749 \\
        \hline
    \end{tabular}
    \caption{Performance of Dinov2-small + 2-scale \textit{Charm} on AADB dataset with different scaling factors. $p'$ and $f$ represent the initial patch size and the downscaling factor, respectively. The patch size $p$ is set to 14 to match the patch size of the Dinov2-small patch encoding module. Increasing $f$ negatively affects performance due to increased information loss during the downscaling of unselected regions.}
    \label{tab:scaling_factor}
\end{table}

\section{The effect of varying number of tokens in Table 3}
We evaluated the standard tokenizer with 384×384 input images (vs.~224×224 used in Table 3), producing 729 tokens — comparable to Charm’s 768 on the Tad66k dataset. Using the Dinov2-small backbone, this setting achieved PLCC/SRCC/ACC of 0.429/0.404/0.653 on TAD66k, while Charm reached 0.488/0.458/0.794.
Charm's strong IAA performance is therefore not due to a higher number of tokens but rather its ability to preserve aspect ratio, composition, and high-resolution details, along with its patch selection strategy that alleviates overfitting.

\section{Downscaling factor}
\label{appendix: scaling factor}
Figure 2 illustrates 2-scale \textit{Charm}, where important areas of the image are further tokenized using a patch size of $p$ while others are downscaled to $p$. As described in Section 3.2.1, the amount of downscaling is defined by $f$. As shown in Table \ref{tab:scaling_factor}, a large $f$ negatively impacts model performance due to increased information loss from downscaling unselected regions.

\section{Integrating Charm with the Swin transformer}
\label{appendix: integration_with_swin}
\textit{Charm} is incompatible with the Swin transformer \cite{liu2021swin} due to Swin's reliance on relative position embeddings and patch merging to capture the hierarchy. Swin transformer requires a specific token order and a fixed grid of patches, which are not guaranteed by \textit{Charm}'s tokenization process. 
While integrating \textit{Charm} with the Swin Transformer presents challenges, it remains a promising direction for future research.

% \section{Model Output Examples}
% Figure \ref{fig:examples} visualizes examples of aesthetic score predictions by our model (Dinov2-small + Charm) compared to the standard Dinov2-small model.

% \begin{figure*}
%     \centering
%     \includegraphics[width=1\linewidth]{figures/examples.jpg}
%     \caption{Examples of the aesthetic scores predicted using our approach (Dinov2-small + Charm) and the standard approach (Dinov2-small). Black, green, and red indicate the ground truth, our approach predictions, and original Dinov2 outputs, respectively. Images are sorted in descending order based on their mean absolute error.}
%     \label{fig:examples}
% \end{figure*}

% WARNING: do not forget to delete the supplementary pages from your submission 

\end{document}